\documentclass[sigconf]{acmart}

\usepackage{comment}
\usepackage{algorithm}
\usepackage{algcompatible}
\usepackage{algpseudocodex}
\usepackage{amsmath,bm}

\newtheorem{assumption}{Assumption}

\AtBeginDocument{%
  }

\setcopyright{acmlicensed}
\copyrightyear{2025}
\acmYear{2025}
\acmDOI{XXXXXXX.XXXXXXX}
\acmConference[MobiHoc'25]{To Theoretically Understand Transformer-Based In-Context Learning for Optimizing CSMA}{Oct 27--30,
  2025}{Houston, TX}
\acmISBN{978-1-4503-XXXX-X/2018/06}

\begin{document}

\title{To Theoretically Understand Transformer-Based In-Context Learning for Optimizing CSMA}





\author{Shugang Hao}
 \authornote{Both authors contributed equally to this research. This work was conducted while Hongbo Li was working at SUTD.}
\affiliation{%
  \institution{Singapore University of Technology and Design (SUTD)}
  \city{Singapore}
  \country{Singapore}}
  \email{shugang\_hao@sutd.edu.sg}

\author{Hongbo Li}
\authornotemark[1]
\affiliation{%
  \institution{The Ohio State University}
  \city{Columbus}
  \country{United States}}
  \email{li.15242@osu.edu}

\author{Lingjie Duan}
 \authornote{Corresponding author.}
\affiliation{%
  \institution{SUTD (current); moving to Hong Kong University of Science and Technology (Guangzhou) in late 2025}
  \city{}
  \country{}}
  \email{lingjieduan@hkust-gz.edu.cn} 






\renewcommand{\shortauthors}{Hao et al.}


\thanks{This work is supported in part by SUTD Kickstarter Initiative (SKI) Grant with no. SKI 2021\_04\_07; and in part by the Joint SMU-SUTD Grant with no. 22-LKCSB-SMU-053.}

\begin{abstract}
The binary exponential backoff scheme is widely used in WiFi 7 and still incurs poor throughput performance under dynamic channel environments. Recent model-based approaches (e.g., non-persistent and $p$-persistent CSMA) simply optimize backoff strategies under a known and fixed node density, still leading to a large throughput loss due to inaccurate node density estimation. This paper is the first to propose LLM  transformer-based in-context learning (ICL) theory for optimizing channel access. We design a transformer-based ICL optimizer to pre-collect collision-threshold data examples and a query collision case. They are constructed as a prompt as the input for the transformer to learn the pattern, which then generates a predicted contention window threshold (CWT). To train the transformer for effective ICL, we develop an efficient algorithm and guarantee a near-optimal CWT prediction within limited training steps. As it may be hard to gather perfect data examples for ICL in practice, we further extend to allow erroneous data input in the prompt. We prove that our optimizer maintains minimal prediction and throughput deviations from the optimal values. Experimental results on NS-3 further demonstrate our approach's fast convergence and near-optimal throughput over existing model-based and DRL-based approaches under unknown node densities.  
 

\end{abstract}



\maketitle

\section{Introduction}

Even with enhanced throughput and low latency in WiFi 7, the heavy contention in unlicensed bands can still trigger severe collisions when many devices co-use the same channel in dynamic network environments (e.g., \cite{guo2023exploiting}, \cite{cordeschi2024optimal}, \cite{dai2024theoretical}). To reduce collisions, the binary exponential backoff (BEB) scheme widely in use (e.g., WiFi 6 \cite{khorov2018tutorial} and WiFi 7 \cite{deng2020ieee}) asks each device or node to wait for a random period whenever a collision occurs, where the contention window threshold (CWT) is doubled after each collision. In dynamic channel environments, it is hard to predict the node density in BEB for determining CWTs, resulting in degraded throughput performance. 


In the CSMA literature, \cite{jiang2009distributed} proposes an adaptive CSMA scheduling algorithm for throughput maximization, assuming no collision between conflicting links. To handle collisions, recent model-based approaches (e.g., non-persistent and $p$-persistent CSMA   \cite{cordeschi2024optimal}, \cite{dai2022theoretical,bianchi2000performance,singh2021adaptive,gao2022aloha}) optimize the backoff strategies under a known node density assumption. They derive the throughput formulation in closed forms for maximization and solve the optimal contention window thresholds in terms of the node density. However, this assumption no longer holds under dynamic channel environments with unknown or varying node densities, leading to a large throughput loss due to an inaccurate estimation of the node density. 


There are also model-free studies proposing deep reinforcement learning (DRL) based approaches for unknown environments in the recent CSMA literature (e.g., \cite{wydmanski2021contention,yan2024deep,lee2024dynamic}). \cite{wydmanski2021contention} proposes a DRL approach to dynamically adjust contention window size based
on turn-around-time measurement of channel status in IEEE 802.11ax networks. \cite{yan2024deep} presents a multi-device distributed DRL framework that intelligently tunes the contention window in IEEE 802.11bn networks to minimize tail latency while maintaining throughput. However, such approaches need to retrain from scratch once the channel environment changes, hard to implement on resource-constrained wireless devices. Though  \cite{lee2024dynamic} proposes a soft actor-critic based approach for robust performance
in response to environmental changes, it still inherits heavy training and inference costs. Thanks to the recent success of AI/LLMs (e.g., \cite{li2025theory},\cite{li2025theory2}), a new idea  arises:
\begin{itemize}
    \item \textit{Q1. How to leverage an LLM for optimizing channel access?}
\end{itemize}



In this paper, we study the theory of LLM transformer-based in-context learning (ICL) for optimizing channel access. ICL refers to the ability of a pretrained transformer to learn a new task simply by conditioning on a few input–output examples in its prompt, leveraging learned patterns and attention mechanisms without fine-tuning (e.g., \cite{dong2024survey}, \cite{rubin2022learning}, \cite{min2022rethinking}). Unlike the costly DRL-based approaches, ICL operates through on-the-fly inference once pre-trained and is widely adaptable to many complex tasks like mathematical reasoning problems (e.g., \cite{liu2024makes}, \cite{lumathvista}). Recently, the communication society has explored the direction of introducing ICL to network design, such as transmission power allocation (e.g., \cite{lee2024llm}, \cite{zhou2024large}), network deployment (e.g., \cite{boateng2024survey}, \cite{sevim2024large},\cite{hao2025RLHF},\cite{hao2025CL}), and network detection (e.g., \cite{song2024neuromorphic}, \cite{zhang2024large}). However, all these studies are empirical, lacking a theoretical analysis with performance guarantees. 

Even in the ICL literature of AI society, existing studies mainly propose frameworks with empirical performance evaluations (e.g., \cite{min2022metaicl,zhang2023makes,bertsch2025context}), lacking theoretical guarantees or insights to guide the design of our ICL-based channel access. There are very few studies on analyzing and providing theoretical guarantees of transformer-based ICL prediction (e.g.,  \cite{zhang2024trained,huang2024context,li2024nonlinear,li2025provable}). \cite{zhang2024trained} investigates
the training dynamics of linear transformers for ICL. \cite{huang2024context} gives convergence guarantees of ICL based on a one-layer transformer with softmax attention and linear mapping functions.  \cite{li2024nonlinear} further analyzes the convergence bounds for binary classification problems. However, all these studies consider either binary outputs or a linear mapping function. Recently, \cite{li2025provable} theoretically generalizes ICL to learn non-linear regression tasks. However, in practical CSMA design, the mapping from collision parameters to the contention window threshold (CWT) is more complex. Moreover, CWTs are defined over a set of integers rather than a simple binary domain. Therefore, the applicability of ICL to real-world CSMA scenarios remains unclear.  Our second question thus arises:
\begin{itemize}
    \item \textit{Q2. How to provide throughput performance guarantee for ICL optimizer design in optimizing channel access?} 
\end{itemize}

There are two technical challenges to design a transformer-based ICL optimizer for channel access. Firstly, ICL involves a non-linear transformer architecture with a softmax operation and its prediction is highly non-convex in the transformer parameters, making it challenging to optimize for a small prediction loss. Secondly, channel environment changes from time to time, requiring our algorithm robust to unknown node densities.

We summarize our key novelty and main results below.
\begin{itemize}
    \item \textit{New theory of transformer-based in-context learning for optimizing channel access:} To our best knowledge, we are the first to propose LLM transformer-based in-context learning (ICL) theory for optimizing channel access. Unlike model-based approaches (e.g., non-persistent and $p$-persistent CSMA) assuming a known and fixed contention node density, we aim for an analytical study not requiring the node density. 
    We provide new analytical insights into \textit{how a system can leverage transformer-based ICL to optimize channel throughput}. 

    \item \emph{Transformer-based ICL design with prediction and throughput performance guarantee:} 
    We propose a transformer-based ICL approach to predict the optimal CWT for any collision case given several pre-collected collision-threshold data examples. We construct these examples and a query collision case as a prompt input to the transformer, which learns the pattern to generate a predicted CWT. We develop an efficient algorithm to train the transformer for effective ICL. Despite the highly non-convex loss objective, our algorithm guarantees a near-optimal prediction within limited training steps, ensuring a converged throughput loss.
    

    
    \item \textit{Practical extension to erroneous data input:} As it may be hard to gather perfect data examples in practice, we further extend to allow erroneous data input in the prompt for ICL. We still manage to prove that the ICL prediction loss approaches the optimum with limited throughput loss. Experimental results on NS-3 further demonstrate our approach's fast convergence and near-optimal throughput over existing approaches under unknown node densities. 
\end{itemize}

The rest of this paper is organized as follows. Section~\ref{S2} introduces the system model and the throughput optimization problem. Section~\ref{S3} discusses the model-based and DRL-based approaches in the CSMA literature as benchmarks. Section~\ref{S4} details our transformer-based ICL approach design for optimizing channel access. Section~\ref{S5} illustrates the analysis of our approach and gives theoretical guarantees. Section~\ref{S6} extends to consider the erroneous prompt case. Section~\ref{S7} conducts experiments to verify our theoretical results. Section~\ref{S8} finally concludes the paper. 


\begin{figure}
    \centering
    \includegraphics[width=0.75\linewidth]{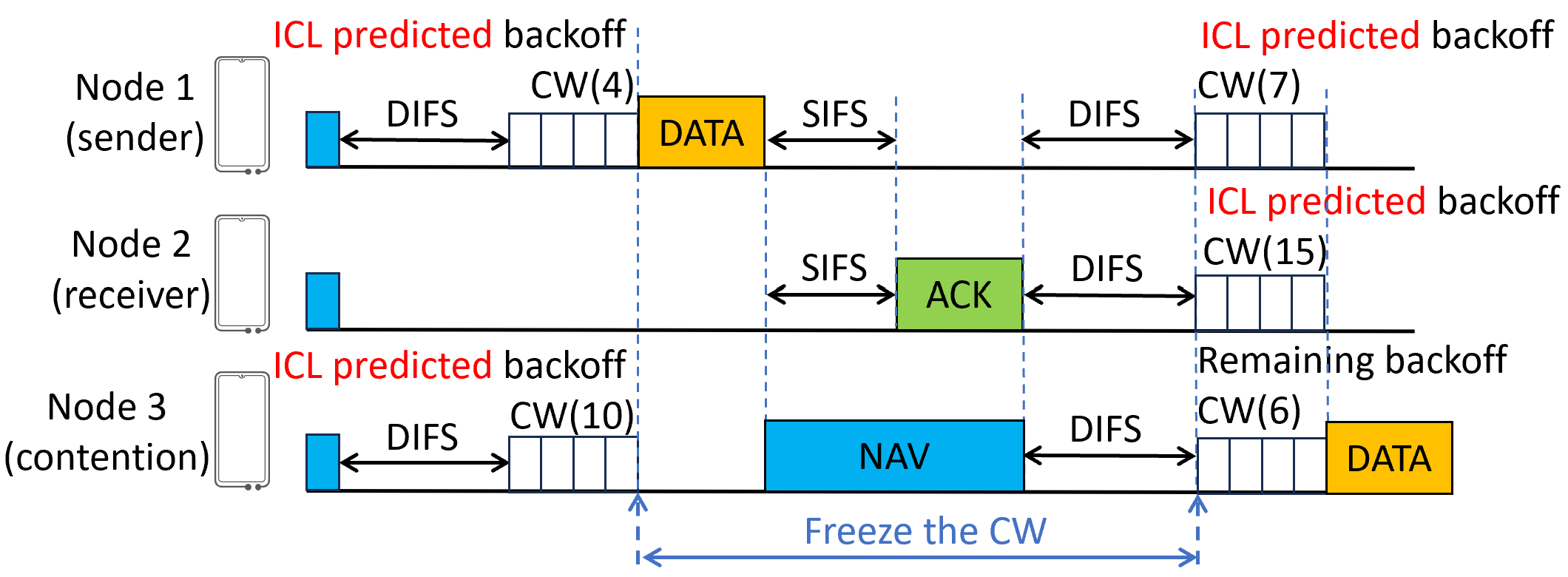}
    \caption{An illustration of our ICL-based NP-CSMA for channel access and collision avoidance, where node 1 sends data to node 2 with node 3 contending on the channel. Our ICL design is illustrated later in Section~\ref{S4}.}
    \label{fig0}
\end{figure}

\section{System Model and Problem Formulation}\label{S2}

In this section, we first introduce our system model based on a typical NP-CSMA. Then, we formulate the throughput maximization problem for further analysis in Sections~\ref{S3}-\ref{S6}.


We follow the renowned distributed coordination function (DCF) for multiple contention nodes' channel access and collision avoidance, which has been widely deployed in practical 802.11 protocols (e.g., the latest WiFi-7 \cite{deng2020ieee}). Based on DCF, we consider a general slotted NP-CSMA with time-varying contention nodes $N(t) \leq \bar{N}$ as follows (also shown in Fig.~\ref{fig0}):

i) The system operates the NP-CSMA in a time-slotted way, where each node's transmission decision is only made at the beginning of each time slot $t \in [T]:=\{1, \cdots, T\}$. It pre-determines the backoff strategy $\{(k, W_k)\}_{k=0}^K$ for all the nodes, where $K$ denotes the maximum collision number for each packet transmission, $k\in\{0, \cdots, K\}$ denotes the collision number since the last successful transmission and $W_k$ is a positive integer to represent the contention window threshold for the collision number $k$ with $W_0<\cdots<W_K\leq \bar{W}$. Note that the above backoff strategy is more general than the BEB scheme widely used in 802.11 protocols, a special case with contention window threshold $W_k = 2^k W_0$.

ii) Each node $i\in[N(t)]$ senses the co-used channel whenever it has a data packet to transmit in a time slot $t$. If the channel is sensed as idle for a distributed interframe space (DIFS), it uses the initial CWT $W_0$ to generate a random integer timer between 0 and $W_0-1$ to backoff before transmission. Otherwise, it does not transmit the packet and keeps sensing until the channel is idle for a DIFS.

iii) Whenever a collision happens, it draws a random integer timer between $0$ and $W_k-1$ according to the current collision number $k$ to backoff. It then counts down the timer and transmits until the timer reduces to 0.

The objective is to find the optimal contention window thresholds $\{W_k^*\}_{k=0}^K$ to maximize the throughput for successful packet transmissions. Following the CSMA literature (e.g., \cite{bianchi2000performance}, \cite{wong2011analysis}, \cite{cordeschi2024optimal}), we define the throughput $U$ as the fraction of time that the channel is used to successfully transmit a packet:
\begin{align}\label{eq0}
    U = \frac{\mathbb{E}[\text{A successful packet transmission per slot}]}{\mathbb{E}[\text{Length of a slot time}]}.
\end{align}
The corresponding throughput maximization problem is thus
\begin{align*}
    \max_{\{W_k\}_{k=0}^K} U(\{W_k\}_{k=0}^K) \ \text{in} \ \eqref{eq0}.
\end{align*}

To maximize the above throughput $U$, model-based approaches (e.g., \cite{dai2022theoretical,cordeschi2024optimal,bianchi2000performance,singh2021adaptive,gao2022aloha}) rely on the exact formulation of $U$ for solving the optimal contention window thresholds. They require the exact knowledge of the packet transmission probability and the successful transmission probability, which are difficult to obtain in dynamic network environments. 



For an improved design of ICL for channel access, a converged loss of our approach's throughput $\hat{U}$ from that at the optimum $U^*$ should be guaranteed within a limited number of training steps. In the rest of this work, for a vector $w$, we let $\|w\|$ denote its $\ell$-$2$ norm. For some positive constant $c_1$ and $c_2$, we define $x=\Theta(y)$ if $c_1|y|<x<c_2|y|$ and $x=\mathcal{O}(y)$ if $x<c_1|y|$. Let $\mathbb{N}$ denote the set of natural numbers.


\section{Benchmarks: Model-Based \& DRL-Based Approaches}\label{S3}

In this section, we first introduce a model-based approach as a benchmark and prove its inefficiency in adapting to unknown node densities. Then, we discuss the DRL-based approaches for further comparison in Section~\ref{S7}.



In the CSMA literature, \cite{jiang2009distributed} focuses on throughput maximization by assuming no collision between conflicting links. To handle collisions, recent model-based approaches (e.g., \cite{cordeschi2024optimal}, \cite{dai2022theoretical,bianchi2000performance,singh2021adaptive,gao2022aloha}) assume a fixed and known node density $N$. 
 They derive the exact formulation of throughput to determine the optimal CWTs. In particular, \cite{bianchi2000performance} constructs a two-dimensional Markov chain to model each node's state transition. 
According to states' transition probabilities and stationary conditions, \cite{bianchi2000performance} obtains that stationary probability $\tau$ that a node transmits in a generic time slot is the unique solution to 
\begin{align}\label{eq1}
    \tau = \frac{2}{(1-p)\sum_{k=0}^{K-1} p^k W_k + p^K W_K + 1},
\end{align}
where $p= 1 - (1-\tau)^{N-1}$ denotes the constant and independent collision probability. 
Based on the transmission probability $\tau$ and the known node density $N$, \cite{bianchi2000performance} formulates throughput $U(\tau)$ as a function of $\tau$:
 \begin{align}\label{eq6}
    U(\tau) = 
    \frac{N \tau (1-\tau)^{N-1} T_P}{(1-\tau)^N T_\sigma +N \tau (1-\tau)^{N-1} (T_s-T_c) +  (1 - (1-\tau)^N) T_c}, 
\end{align}
where $T_P$ is the packet payload time, $T_\sigma$ is the length of an empty slot time, $T_s$ is the average time that the channel is sensed busy because of a successful transmission, and $T_c$ is the average time the channel is sensed busy during a collision.





To maximize the throughput $U(\tau)$ in \eqref{eq6}, \cite{bianchi2000performance} optimizes the contention window thresholds $\{W_k\}_{k=0}^K$ in \eqref{eq1} for reaching the optimal packet transmission probability $\tau^*$ given the knowledge of the node density $N$. Unfortunately, we have the following under a dynamic and unknown node density. 
\begin{theorem}\label{L4}
Suppose that $\bar{W} \geq 2\bar{N}-1$. In a dynamic channel environment with an unknown node density $N(t)$, the model-based approach with inaccurate estimation $\hat{N}(t)$ of $N(t)$ leads to a large throughput loss
\begin{align*}
U(\tau(N(t))) - U(\tau(\hat{N}(t))) \geq \Theta \bigg(\frac{T_\sigma}{\bar{N}K^2\bar{W}^3T_c^2}\bigg) |N(t) - \hat{N}(t)|.
\end{align*}
\end{theorem}

The proof is given in Appendix~\ref{A1}. Theorem~\ref{L4} indicates that the benchmark leads to a certain throughput loss due to the inaccurate estimation of the node density $N(t)$. 
In highly dynamic channel environments, the gap between the actual $N(t)$ and the estimated $\hat{N}(t)$ is large, leading to a large throughput loss. This motivates us to further develop an approach well adapting to unknown node densities.

There are model-free studies proposing deep reinforcement learning (DRL) based approaches for unknown environments in the recent CSMA literature (e.g., \cite{wydmanski2021contention,yan2024deep,lee2024dynamic}). \cite{wydmanski2021contention} proposes a DRL approach to dynamically adjust contention window size based
on turn-around-time measurement of channel status in IEEE 802.11ax networks. \cite{yan2024deep} presents a multi-device distributed DRL framework that intelligently tunes the contention window in IEEE 802.11bn networks to minimize tail latency while maintaining throughput. However, such approaches need to retrain from scratch once the channel environment changes, hard to implement on resource-constrained wireless devices. Though  \cite{lee2024dynamic} proposes a soft actor-critic based approach for robust performance
in response to environmental changes, it still inherits the heavy training and inference costs.
Later in Section~\ref{S7.4}, we run experiments to show our approach's much faster convergence than soft actor-critic. 

According to the above analysis in Section~\ref{S3}, we are well motivated to propose a novel and efficient transformer-based ICL  optimizer in Section~\ref{S4}, 
which only involves a few collision-threshold examples and does not need the knowledge of the unknown and varying node density.

\section{Transformer-Based ICL for Optimizing Channel Access}\label{S4}

\begin{figure}
    \centering
    \includegraphics[width=0.72\linewidth]{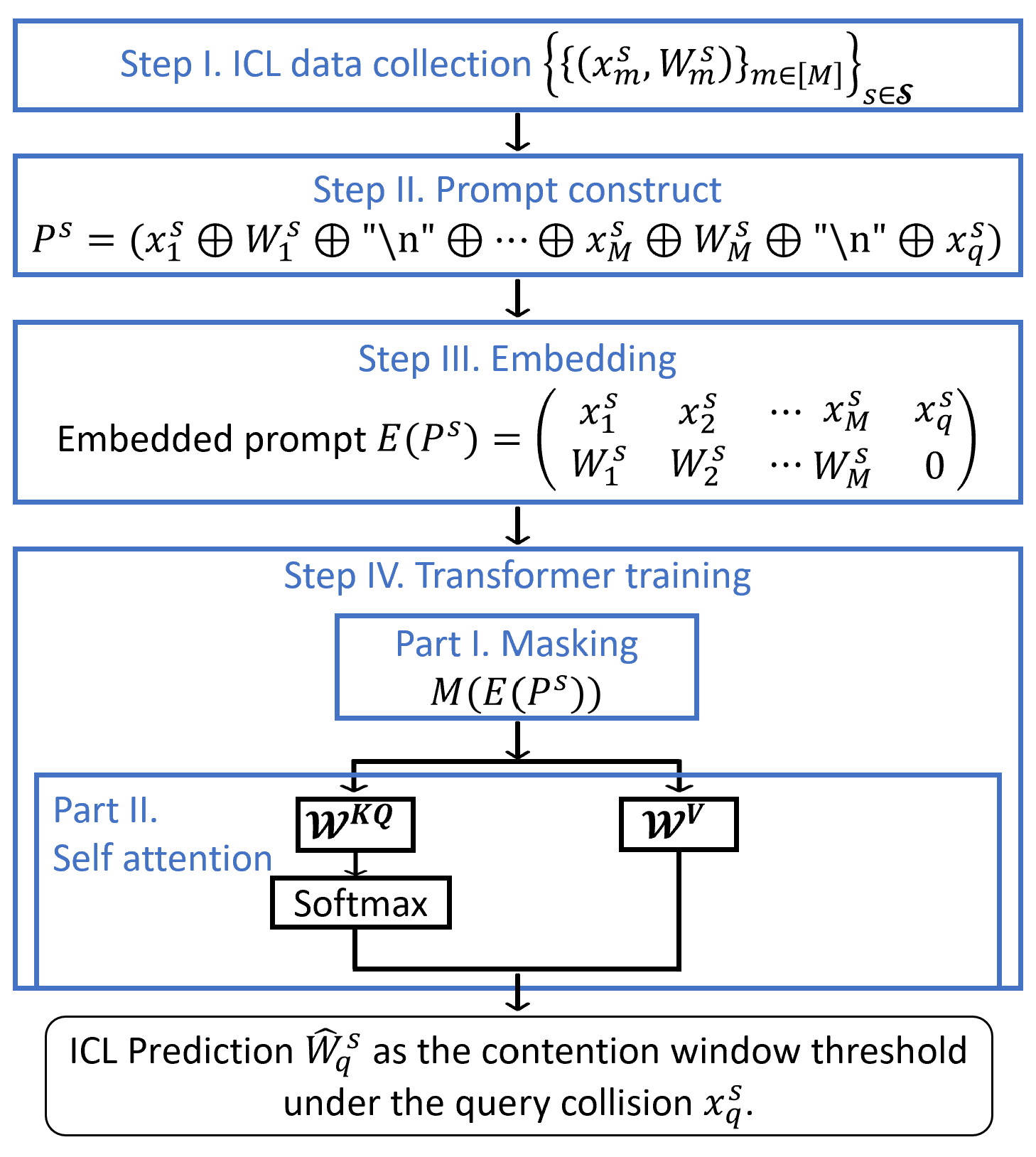}
    \caption{The overview of our transformer-based ICL optimizer in four steps: ICL data collection, prompt construction, embedding, and transformer training. First, the system collects $M$ prior data points $\{(x_m^s, W_m^s)_{m\in[M]}\}$ and a query $x_q^s$ as a new input with an unknown $W_q^s$ in node density case $s \in \mathcal{S}$. Secondly, the system constructs prompt $P^s$ using data points $\{(x_m^s, W_m^s)_{m\in[M]}\}$ and the query pair $(x_q^s,0)$ for each $s$. Thirdly, each prompt $P^s$ is embedded as $E(P^s)$, which is further masked as $\mathcal{M}(E(P^s))$ to prevent the query input from attending to itself in the transformer training. Finally, $\mathcal{M}(E(P^s))$ is multiplied by the $\mathcal{W}^{KQ}$ and $\mathcal{W}^V$ matrices with the self-attention mechanism to obtain the final prediction $\hat{W}_q^s$ in \eqref{eqW} as the contention window threshold under the query collision $x_q^s$. Our goal is to determine a near-optimal prediction $\hat{W}_q^s$ for the new query collision $x_q^s$ based on the prior data points $\{(x_m^s, W_m^s)_{m\in[M]}\}$.}
    \label{fig4-1}
\end{figure}

As shown in Figure~\ref{fig4-1}, our transformer-based ICL optimizer contains four steps: ICL data collection, prompt construction, embedding, and transformer training. Step I prepares the collision-threshold examples and the query collision case for prompt construction. Steps II and III construct prompts in efficient forms as the transformer input.  Step IV performs the transformer's self-attention mechanism to infer the optimal contention window. We detail our design of each step in the following.

\subsection{Step I: ICL Data Collection}\label{S4.1}

We define $x\in\mathbb{R}^d$ as the feature vector containing a set of $d$-dimensional collision parameters (e.g., the current collision number $k$, payload transmission time $T_P$, successful transmission time $T_s$, collision transmission time $T_c$, etc.). In practice, due to propagation delay variations, hardware clock inaccuracies, and processing jitter, these data related to packet transmission are noisy to fluctuate from time to time. 
Since the space of collision parameters is finite, we summarize all possible feature vectors $x_i$ into a collision set $\mathcal{X}:=\{x_i\in\mathbb{R}^d| i \in [I]\}$, where $\|x_i-x_{i'}\|=\Theta(\Delta)$ with feature vector gap $\Delta= \Theta(1)$ for any $i\neq i'$.

In a dynamic channel environment, since the node density changes dynamically over time, we use $\mathcal{S}\in \mathbb{R}^{d_s}$ to summarize all the possible node density environments.
To ensure effective transformer training and adaptation to the dynamic node density environments, the system must sample multiple data pairs under different node densities.
Consequently, for each node density environment $s\in\mathcal{S}$, we collect $M$ prior data points for transformer training, where we let $x_m^s$ denote each parameter for $m\in[M]$. We suppose $x_m^s$ is a noisy version of some $x_i\in\mathcal{X}$, satisfying $\|x_m^s-x_{i}\|=\mathcal{O}(\Delta)$. For simplicity, we assume $x_m^s=x_k$ in the later analysis, such that each $x_m^s$ is randomly sampled from $\mathcal{X}$ with probability $p_k=\Theta(\frac{1}{K})$.
For each $x_m^s$ of node density environment $s$, we let $W_m^s$ denote its corresponding optimal contention window threshold under the current mapping function, defined as $f^s: \mathcal{X}\rightarrow \mathbb{N}$. In other words, we have $W_m^s=f^s(x_m^s)$.
Note that the mapping functions $f^s$ are unknown and vary with node density $s$.

For each feature vector $x_m^s$, we need to obtain its corresponding optimal $W_m^s$ later for prompt construction and transformer training. We use a branch-and-bound algorithm (e.g., \cite{morrison2016branch}) to solve the optimal probability $\tau^*$ that maximizes throughput $U$ in \eqref{eq6} under a node density. Given $\tau^*$, we then apply mixed-integer linear programming techniques (MILP), such as golden section search \cite{chang2009n} and parabolic interpolation \cite{heath2002introductory}, to solve for the optimal contention window thresholds $\{W_m^s\}_{m=1}^M$ in \eqref{eq1}. 
After successfully forming $M$ data pairs $\{\{(x_m^s, W_{m}^s)\}_{m \in [M]}\}_{s\in \mathcal{S}}$ as prior training data, we sample a query $x_q^s$ as a new input with an unknown $W_q^s$ to be decided by our ICL optimizer.
In Section~\ref{S6}, we relax this assumption of known optimal $W_{m}^s$ and analyze the impact of erroneous data collection about contention window thresholds.

\subsection{Steps II \& III: Prompt Construction and Embedding}\label{S4.2}

After obtaining $M$ data pairs $\{(x_m^s, W_{m}^s)\}_{m \in [M]}$ and the new query $x_q^s$ by Step I, the system constructs each prompt 
\begin{align}\label{prompt}
    P^s \!\!=\!\! x_1^s \oplus W_{1}^s \oplus ``\texttt{\textbackslash n}" \oplus \cdots \oplus x_M^s \oplus W_{M}^s \oplus ``\texttt{\textbackslash n}" \oplus x_{q}^s, 
\end{align}
where $s \in \mathcal{S}$, $\oplus$ denotes the string concatenation operator, $``\verb|\n|"$ denotes a special delimiter token to distinguish data input. The last term $x_{q}^s$ serves as the query of current collision parameter vector for predicting the optimal contention window threshold $W_q^s$, referred to as the query token. 
Since all examples within a prompt $P^s$ correspond to the same node density, each collision pair of $\{(x_m^s, W_m^s)\}_{m\in[M]}$ and the query pair $(x_q,W_q^s)$ follow the same mapping $f^s \in \mathcal{F}$, satisfying $W_m^s = f^s(x_m^s)$ for any $m \in [M]$ and $W_q^s=f^s(x_q^s)$. One prompt example is $P^s$=(1,8;3,32;5,128;2) with $d=1$.

Given each prompt $P^s$ in \eqref{prompt}, we follow a natural token embedding in the ICL literature (e.g., \cite{huang2024context,li2025provable}) to construct each column $m \in [M]$ as $\begin{pmatrix}
    x_m^s\\W_m^s
\end{pmatrix}$ and the last column as $\begin{pmatrix}
    x_q^s\\0
\end{pmatrix}$.
Thus, we obtain the embedding matrix of each $P^s$ below:
\begin{align}
E(P^s) = \begin{pmatrix}
x_{1}^s    & \cdots & x_{M}^s       & x_{q}^s \\
W_{1}^s      & \cdots & W_{M}^s & 0
\end{pmatrix}
\in \mathbb{R}^{(d+1) \times (M+1)}.\label{E(Ps)}
\end{align}
Next, we use this embedding $E(P^s)$ to train the transformer.

\subsection{Step IV: Transformer Training}\label{S4.3}

 Similar to the existing ICL literature (e.g., \cite{huang2024context,zhang2024trained,li2025provable}), we consider a simple but fundamental one-layer transformer, which contains a masking part and a self-attention part as in Figure~\ref{fig4-1}. While simple, this facilitates our later theoretical analysis in Section~\ref{S5}, which demonstrates its effectiveness in optimizing channel access and shows enough advantages over the benchmarks of Section~\ref{S3}.

To improve training efficiency, the system applies a masking operation to each embedding matrix $E(P^s)$ in \eqref{E(Ps)}, producing $\mathcal{M}(E(P^s))$, which removes the last column to prevent the query input from attending to itself during training. The transformer then performs self-attention on the masked embedding $\mathcal{M}(E(P^s))$.
We define the self-attention mechanism below.
\begin{definition}
    A self-attention (SA) layer in the single-head case consists a key matrix $\mathcal{W}^{K} \in \mathbb{R}^{{(d+1) \times (d+1)}}$, a query
matrix $\mathcal{W}^Q \in \mathbb{R}^{{(d+1) \times (d+1)}}$, and a value matrix $\mathcal{W}^V \in \mathbb{R}^{{(d+1) \times (d+1)}}$. Given an embedding $E$ of a prompt $P$, the self-attention mechanism outputs
\begin{align*}
    F_{SA}\bigl(E; \mathcal{W}^{K}, \mathcal{W}^Q, \mathcal{W}^V\bigr)
\!\!=\!\! \mathcal{W}^V E \!\cdot\! \mathrm{softmax}\!\Bigl(\bigl(\mathcal{W}^{K} E\bigr)^{\!\top} \mathcal{W}^Q E\Bigr),
\end{align*}
where the softmax function is $\operatorname{softmax}(\bm{z}_i)= \frac{\exp(\bm{z}_i)}{\sum_{j}\exp(\bm{z}_j)}$ with $\bm{z}_i$ denoting the $i$-th element of vector $\bm{z}$.
\end{definition}

We normalize the value matrix $\mathcal{W}^{V}$ to represent equal contribution from each collision-threshold pair in a prompt. Further, we consolidate the query and key matrices into one matrix as $\mathcal{W}^{KQ} \in \mathbb{R}^{d \times d}$ in the following forms: 
\begin{align}\label{QKV}
    \mathcal{W}^{V} = 
\begin{pmatrix}
0_{d \times d} & 0_{d} \\
0_{d}^{\top} & 1
\end{pmatrix},
\quad
\mathcal{W}^{KQ} =
\begin{pmatrix}
Q & 0_{d} \\
0_{d}^{\top} & 0
\end{pmatrix}.
\end{align}
Note that the consolidation operation on the $\mathcal{W}^{KQ}$ matrix does not change the softmax input in $F_{SA}$. For ease of exposition, we use the notation $\theta = (1, Q)$ to represent all the transformer parameters for simplifying the transformer training analysis later in Section~\ref{S5}. Next, we are ready to give the self-attention mechanism in the parameter $\theta$ as follows:
\begin{align}\label{SA}
    &F_{\text{SA}}(E(P^s); \theta) \nonumber \\
    =&  \mathcal{M}(E^{W}(P^s)) \cdot  \text{softmax} \left( \mathcal{M}(E^{x}(P^s))^{\top} Q E^{x}(P^s) \right),
\end{align}
where $E^{x}(P^s)$ and $E^{W}(P^s)$ denote the first $d$ rows and the last row of $E(P^s)$, respectively. 
The ICL prediction for the query collision  $x_q^s$ is the last entry of $F_{\text{SA}}$ as follows:
\begin{align}\label{eqW}
    \hat{W}_{q}^s = \hat{W}_{q}^s(E(P^s); \theta) = \left[ F_{\text{SA}}(E(P^s); \theta) \right]_{(M+1)}.
\end{align}

For transformer training under different node densities, we aim to minimize the following squared loss of  prediction error:
\begin{align}
    \mathcal{L}(\theta)=\mathbb{E}_{f^s\in\mathcal{F}}[(\hat{W}_{q}^s-W_q^s )^2],\label{eq7}
\end{align}
where $W_q^s$ is the optimal contention window threshold of $x_q^s$ under $f^s$ derived in Step I. 

\begin{algorithm}[t]
\caption{Gradient descent for transformer training in Step IV of our transformer-based ICL optimizer}
\begin{algorithmic}[1] 
\Require Training loss objective $\mathcal{L}(\theta)$ in \eqref{eq7}, transformer parameter $\theta = (1, Q)$, maximum training round number $\bar{T}$, step size $\eta$, precision error $\epsilon$.

\Ensure Trained transformer parameter $\theta^*$.

\State{\textbf{Initialization}: $Q^{(0)}\leftarrow \textbf{0}_{d \times d}$, $\theta^{(0)} \leftarrow (1, Q^{(0)})$.}

\State{Update the transformer parameter via gradient descent:
\[
\centering
\theta^{(t+1)} = \theta^{(t)} - \eta \cdot \nabla_{\theta} \mathcal{L}(\theta^{(t)}).
\]
}
\State{Record $\theta^* \leftarrow \theta^{(t+1)}$ if $\exists t < \bar{T}$ such that $|| \theta^{(t+1)} - \theta^{(t)} ||_2 \leq \epsilon$. Otherwise, Record $\theta^* \leftarrow \theta^{(\bar{T})}$. }

\end{algorithmic}
\end{algorithm}

Observing $F_{\text{SA}}(E(P^s); \theta)$ in (\ref{SA}), we find that the non-linear softmax function couples the attention weights across all input tokens, making the training loss objective $\mathcal{L}(\theta)$ in \eqref{eq7} highly non-convex and interdependent. Although we have reduced the parameter space to $\theta=(1, Q)$, it is still difficult to explicitly solve the closed-form $\theta^*$ for minimizing $\mathcal{L}(\theta)$ in (\ref{eq7}) with standard techniques in optimization theory. 
Consequently, we aim to propose a simple algorithm to efficiently train the transformer with a convergence guarantee after a limited number of training steps. To achieve this goal, we employ the gradient descent algorithm to optimize the non-convex and high-dimensional loss functions, which offers an efficient and scalable way to deal with our highly non-convex objective and is widely adopted in the machine learning literature (e.g., \cite{huang2024context}, \cite{li2024nonlinear}, \cite{zhang2024trained}). We then summarize details in Algorithm 1.

Although Algorithm 1 provides an efficient way to train the transformer parameters, it remains unclear how the parameter $\theta$ evolves during the training step, how long the training procedure takes at most, and whether the convergence of the final parameter $\theta^*$ can theoretically be guaranteed. We then make a comprehensive theoretical analysis in the next section. 


\section{Performance Analysis of Transformer-Based ICL for Optimizing Channel Access}\label{S5}

\subsection{Convergence Analysis of Algorithm 1}\label{S5.1}

In this section, we aim to prove Algorithm~1's convergence and a limited throughput loss from the optimum. We need to build the connection between any collision example in the prompt and the query collision for a good transformer training and ICL prediction. To achieve this, we first define attention scores to measure how much the new query token $x_q^s$ of $P^s$ in (\ref{E(Ps)}) attends or relates to other input collision parameter vectors $\{x_m^s\}_{m\in[M]}$ in transformer training as in part II of Figure~\ref{fig4-1}. 

\begin{definition}
    Given a prompt $P^s$ in \eqref{prompt} and its embedding $E(P^s)$ in \eqref{E(Ps)}, we define the attention score at time $t$ for self-attention mechanism $F_{SA}$ in \eqref{SA} with parameter $\theta^{(t)}$ as below.
    
a) Given $m \in [M]$, the attention score for the $m$-th collision token $x_m$ is
\begin{align*}
            \mathrm{attn}_m^{(t)}(\theta^{(t)};E^s(P^s)) \!:=\!& \bigg[\text{softmax}\!\!\left( \mathcal{M}(E^{x}(P^s))^{\top} Q E^{x}(P^s) \right)\!\!\bigg]_m \\
            =&
\frac{\exp\!\Bigl(\bigl(E^x_m(P^s)\bigr)^{\!\top}\,Q\bigl(\theta^{(t)}\bigr)E^{x}_{M+1}(P^s)\Bigr)}
     {\displaystyle \sum_{j \in [M]} \exp\!\Bigl(\bigl(E^x_j(P^s)\bigr)^\top\,Q\bigl(\theta^{(t)}\bigr)E^{x}_{M+1}(P^s)\Bigr)}.
        \end{align*}
    
         b) For $k \in [K]$, define $\mathcal{X}_k^s(P^s) \subset [M]$ as the index set of collision input such that $x_m^s = x_k$ for $m \in \mathcal{X}_k^s(P^s)$. Then the attention score for the $k$-th collision token is given by
         \begin{align}
             \mathrm{Attn}_k^{(t)}(\theta^{(t)};E^s(P^s)) := \!\!\!\!
\sum_{m \in \mathcal{X}_k^s(P^s)} \!\!\!\! \mathrm{attn}_m^{(t)}(\theta^{(t)};E^s(P^s)).\label{defeq_Attn}
         \end{align}
\end{definition}

For simplicity, we represent $\mathrm{attn}_m^{(t)}(\theta^{(t)};E^s(P^s))$ as $\mathrm{attn}_m^{(t)}$ and $\mathrm{Attn}_k^{(t)}(\theta^{(t)};E^s(P^s))$ as $\mathrm{Attn}_k^{(t)}$, respectively. We further rewrite $\mathcal{X}_k^s(P^s)$ as $\mathcal{X}_k^s$. Then the ICL prediction $\hat{W}_q^s$ in \eqref{eqW} for the current query token $x_q^s$ can be rewritten as
\begin{align}
    \hat{W}_q^s = \sum_{m \in [M]} \mathrm{attn}_m^{(t)} W_m^s = \sum_{k \in [K]} \mathrm{Attn}_k^{(t)} f^s(x_k).\label{W_output}
\end{align}

Following the ICL literature \cite{li2025provable}, we consider a broader class of non-degenerate $L$-Lipschitz continuous functions:
\begin{assumption}[non-degenerate $L$-Lipschitz \cite{li2025provable}]\label{assump1}
  Each mapping function $f^s$ is $L$-Lipschitz continuous, i.e.,
  \begin{align}
      |f^s(x) - f^s(x')| \leq L\|x-x'\|, \forall \|x-x'\|=\Theta (\delta_0),\label{assump_1}
  \end{align}
  where $L>0$ and $\delta_0=\mathcal{O}(1)$. Moreover, for every $x_k\in\mathcal{X}$, there exists some $x_{k'} \in \mathcal{X}$ with $k'\neq k$, such that
  \begin{align}
      |f^s(x_k) - f^s(x_{k'})| = \Theta(L)\cdot\|x_k - x_{k'}\|.\label{assump_2}
  \end{align} 
\end{assumption}

For notational convenience, we write $f^s(x) \in \mathcal{F}$ if $f^s(x)$ satisfies Assumption~\ref{assump1}.
The global $L$-Lipschitz condition in (\ref{assump_1}) ensures that the function $f^s$ changes smoothly without excessive variation, a common assumption in both theoretical and empirical studies encompassing a wide range of linear and nonlinear mappings. The additional non-degeneracy condition in (\ref{assump_2}) guarantees that $f^s$ maintains sufficient separation between inputs, ensuring that different instances remain distinguishable, which a critical property for enabling learnability in ICL tasks.
There are many typical functions satisfying non-degenrate $L$-Lipschitz in Assumption~\ref{assump1}, such as non-constant linear, exponential, and ReLu functions. 
In Section~\ref{S7}, we relax Assumption~\ref{assump1} in experiments to verify the near-optimal throughput under our optimizer.






Building on the expression in \eqref{W_output}, we follow \cite{li2025provable} to characterize how the attention scores defined in \eqref{defeq_Attn} influence the prediction loss defined in \eqref{eq7} in the following lemma. 
\begin{lemma}[\cite{li2025provable}]\label{lemma:loss}
Given constants $L,\Delta>0$, for any $t\in[T]$, the prediction loss in \eqref{eq7} can be expressed as:
\begin{align}\label{loss_attn}
   \mathcal{L}(\theta)=\frac{1}{2}\sum_{k=1}^K\mathbb{E}\Big[\mathbf{1}\{x_{q}^s=x_k\}\Big(1-\mathrm{Attn}_k^{(t)}\Big)^2\cdot \mathcal{O}(L^2\Delta^2)\Big],
\end{align}
where $\mathbf{1}\{x_{q}^s=x_k\}=1$ is the indicator function that equals 1 if $x_{q}^s = x_k$ and 0 otherwise.
\end{lemma}
\begin{proof}
See Appendix~\ref{A0}.
\end{proof}

Lemma~\ref{lemma:loss} reveals that the prediction loss $\mathcal{L}(\theta)$ depends on the function Lipschitz constant $L$, the feature gap $\delta_0$, and the attention score $\mathrm{Attn}_k^{(t)}$ associated with the true feature of the query token (i.e., $x_q^s=x_k$). 

Based on Lemma~\ref{lemma:loss}, we can analyze the convergence of loss $\mathcal{L}(\theta)$. Define $T^*$ as the convergence time of Algorithm~1. Together with Definition~2, we then establish the following criterion for Algorithm~1's convergence. 
\begin{proposition}\label{prop_convergence}
Given $\delta_0=\mathcal{O}(1)$, our Algorithm~1 achieves convergence if and only if, for any $t>T^*$ with query token $x_{q}^{s}=x_k$, the attention score with respect to $x_k$ satisfies
\begin{align}
    1-\text{Attn}_k^{(t)}=\mathcal{O}(\delta_0).\label{convergence}
\end{align}
\end{proposition}
\begin{proof}
See Appendix~\ref{A2}.
\end{proof}

Intuitively, in the convergence state, if the query token is $x_{q}^{s}=x_k$, the transformer correctly outputs the optimal contention window threshold $f^s(x_k)$ for $x_q^s$ by assigning an attention score $\text{Attn}_k^{(t)}$ close to $1$ in (\ref{W_output}). 
Based on the convergence in attention score $\text{Attn}_k^{(t)}$ in Proposition~\ref{prop_convergence}, we now establish the convergence guarantee for Algorithm~1.
\begin{theorem}\label{T5.3}
If the number of input examples $M$ in a prompt $P^s$ in \eqref{prompt} and the maximum collision number $K$ satisfy $M\geq\text{poly}(K)$, and the gap parameter $L$ of mapping functions satisfies $L=\mathcal{O}(\frac{1}{\Delta})$, then for any $\epsilon\in(0, 1)$, applying Algorithm 1 to train the loss function $\mathcal{L}(\theta)$ in \eqref{eq7} ensures convergence. Specifically, with at most $T^* =\Theta(\frac{K\log(K\epsilon^{-1})}{\eta\delta_0^2 L^2\Delta^2})$ iterations, the prediction loss satisfies $\mathcal{L}\big(\theta^{(T^*)}\big)=\mathcal{O}(\epsilon^2)$.
\end{theorem}
\begin{proof}
See Appendix~\ref{A3}.
\end{proof}

Theorem~\ref{T5.3} states that with a sufficiently large number of data examples $M$, Algorithm 1 ensures convergence of the loss function $\mathcal{L}(\theta)$ in (\ref{eq7}) within a limited number $T^*$ of training steps, allowing the transformer to output a near-optimal $\hat{W}_q^s$ for $x_q^s$.
However, as the maximum collision number $K$ increases, each prompt needs to include more data examples to train, which delays the convergence.
Conversely, the convergence time $T^*$ in Theorem~\ref{T5.3} decreases with both the function gap parameter $L$ and feature vector gap $\Delta$ defined in Assumption~1 since a larger $L$ or $\Delta$ results in a greater gradient, which accelerates the loss function convergence.
Unlike DRL benchmarks, which inherit heavy training and inference costs, our transformer-based ICL optimizer adapts efficiently by focusing on the iteration count needed for convergence, making it well-suited for dynamic node density environments. After training, the inference delay of our $d$-dimensional transformer for a prompt with $n$ tokens is $\mathcal{O}(n^2d)$.

\subsection{Theoretical Guarantee of Throughput Performance}\label{S}

To further check the throughput $U$ in \eqref{eq0} of our transformer-based ICL approach, we define the throughput loss $\Delta U$ as the expected difference between the throughput $U^*$ under the optimal contention window threshold ${W}_{q}^s$ and $\hat{U}$ of the ICL prediction $\hat{W}_{q}^s$ for all the prompts:
\begin{align}\label{eq8}
    \Delta U := U^* - \hat{U} = \mathbb{E}_{f^s\in\mathcal{F}}\big[ U({W}_{q}^s) - U(\hat{W}_{q}^s)\big].
\end{align}


To bound the throughput loss $\Delta U$ in \eqref{eq8} given the training loss of our approach in Theorem~\ref{T5.3}, we further prove the Lipschitz continuity of $U$ according to \eqref{eq1} and \eqref{eq6} as follows.
\begin{lemma}\label{L5.4}
   The throughput $U$ in \eqref{eq6} is $\frac{T_P\bar{N}}{8T_\sigma}$-Lipschitz continuous in each contention window threshold $W_k$: 
    \begin{align*}
        |U(W_k) - U(\hat{W}_k)| \leq \frac{T_P\bar{N}}{8T_\sigma}\cdot |W_k - \hat{W}_k|.
    \end{align*} 
\end{lemma}
\begin{proof}
See Appendix~\ref{A4}.
\end{proof}

According to Theorem~\ref{T5.3} and Lemma~\ref{L5.4}, we are now ready to well bound the throughput loss $\Delta U$ in \eqref{eq8} explicitly.  

\begin{theorem}\label{prop1}
    The throughput loss $\Delta U$ in \eqref{eq8} of our transformer-based ICL approach from the optimum is upper bounded as follows:
    \begin{align*}
        \Delta U \leq  
 \mathcal{O}\bigg(\frac{T_P\bar{N}\epsilon}{8T_\sigma}\bigg).
    \end{align*}
\end{theorem}
\begin{proof}
See Appendix~\ref{A5}.
\end{proof}

Compared with the model-based benchmark's large throughput loss in Theorem~\ref{L4}, Theorem~\ref{prop1} guarantees a limited throughput loss for our transformer-based ICL approach. 
Intuitively, as the maximum node density $\bar{N}$ increases, packet collision may occur more frequently and the system needs to carefully determine the contention window thresholds for reducing collisions to improve the throughput $U$. Accordingly, a small deviation from the optimum contention window thresholds can still lead to a large throughput loss $\Delta U$, resulting in a larger  $\mathcal{O}(\frac{T_P\bar{N}\epsilon}{8T_\sigma})$. 

\section{Practical Extension to Erroneous Data Input}\label{S6}
Recall that in Step I of data collection in Section~\ref{S4.1}, we assume that in each prompt $P^s$ in \eqref{prompt}, each CWT $W_m^s$ is optimal for the collision feature vector $x_m^s$, $m \in [M]$ and $s \in \mathcal{S}$. In practice, it can be difficult for the system to collect the optimal CWT for each collision case. In this section, we remove this assumption and allow each $W_m^s$ to be erroneous for collision $x_m^s$ for further evaluation.

\subsection{Extended System Model of the Erroneous Data Input}\label{S6.1}

The analysis of our transformer-based ICL optimizer in Sections~\ref{S4} and \ref{S5} requires each collision-threshold pair $(x_m^s, W_m^s)$ to follow the same mapping $f^s$ in each prompt $P^s$, $s \in \mathcal{S}$, which no longer holds if $W_m^s$ is erroneous for $x_m^s$. We are then motivated to develop our analysis based on a general LLM, which is pre-trained to learn an arbitrary $f \in \mathcal{D}$ between the collision parameters and the corresponding contention window threshold under another node density. Denote $\mathbb{P}_{f}(y|x)$ as the probability that the LLM generates a contention window threshold $W$ given an input collision $x$ under the mapping $f$, where $x,W$ are taken from a set $\Omega$. 

We consider a challenging case that the mapping $f^* \in \mathcal{D^*}$ in the prompt $P$ to learn 
 under an unknown target node density is not the same as $f$ for pre-training of another node density in general.  In practice, the prompt $P$ containing collision-threshold examples does not resemble inputs that the LLM has been pre-trained on. Thus, either two consecutive example strings $s_1 = x_1 \oplus W_1 \oplus \cdots \oplus x_m \oplus W_m$ and $s_2 = x_{m+1} \oplus W_{m+1}$ from the prompt set $\Omega^*$ with $m \leq M-1$ are approximately independent according to the pretraining mapping as follows:
\begin{align*}
    \alpha \mathbb{P}_{f}(s_2|s_1 \oplus ``\mathrm{\backslash n}") \leq  \mathbb{P}_{f}(s_2) 
    \leq \frac{1}{\alpha} \mathbb{P}_{f}(s_2|s_1 \oplus ``\mathrm{\backslash n}"),
\end{align*}
where $\alpha \in (0, 1]$. To avoid zero likelihood due to the unnatural concatenation of collision-threshold examples in the prompt $P$, we assume that there exists a constant $\beta > 0$ such that for any token $t$ in the prompt set $\Omega^*$, any token $t'$ in the pretaining set $\Omega$ and any mapping $f \in \mathcal{F}$, we have $\mathbb{P}_{f}(t|t') > \beta$. Finally, we consider that there is a positive probability that the pretraining distribution $\mathcal{D}$ generates the mapping $f^*$ of the prompt, i.e., $Pr(f^*|\mathcal{D}) \geq \gamma > 0$. 

To investigate the ICL performance, we model the zero-one loss of the in-context predictor as follows (e.g., \cite{wies2023learnability}):
\begin{align}\label{eq9}
    \mathcal{L} := \mathbb{E}_{x,W\sim \mathcal{D^*}} [\mathbf{1}(\arg\max_{W'} \mathbb{P}_{\theta}(P \oplus W') \ne W)].
\end{align}
Note that $\mathcal{L}$ in \eqref{eq9} is similar to that in \eqref{eq7} to capture the ICL prediction loss to the optimal contention window thresholds.

\subsection{Robustness Analysis to Erroneous Data Inputs}\label{S6.2}

Denote $\ell$ as the length of each collision input $x$. In the following, we first introduce a lemma for further analysis.

\begin{lemma}\label{L6-2}
   Suppose that the minimum KL-divergence between our ICL mapping $f$ and the ground-truth $f^*$ satisfies $\min_{f} KL(\mathbb{P}_{f}, \mathbb{P}_{f^*})$$> -8\ln(\alpha \beta)$. If the number of in-context examples $M$ satisfies
   \begin{align*}
     M \geq  \max\bigg\{\frac{(\ln \frac{1}{q}) (16\ell^2)(\ln^2\beta)}{KL^2(\mathbb{P}_{f}, \mathbb{P}_{f^*})}, \frac{-2\ln(\frac{\mathbb{P}_{\mathcal{D^*}}(W|x) - \mathbb{P}_{\mathcal{D^*}}(\hat{W}|x)}{5\alpha^{-2}\beta^{-T}\gamma^{-1}})}{\min_{f} KL(\mathbb{P}_{f}, \mathbb{P}_{f^*})+8\ln(\alpha \beta)}\bigg\},
   \end{align*}
   $\mathbb{P}_{\mathcal{D^*}}(W|x) - \mathbb{P}_{\mathcal{D^*}}(\hat{W}|x) > 0$ and any $q \in (0, 1)$ for every collision case $x$ and two window candidates $W$, $\hat{W}$, we have
   \begin{align}
    &Pr\bigg( \frac{\mathbb{P}_{\mathcal{D^*}}(W|x) - \mathbb{P}_{\mathcal{D^*}}(\hat{W}|x)}{2} - \big(\mathbb{P}_{\mathcal{D}}(W|P) - \mathbb{P}_{\mathcal{D}}(\hat{W}|P)\big)  \nonumber \\
    &\ \ \ \ \ \ <  1 - \alpha^2\bigg)
    \geq 1 - q. \label{e2}
\end{align}
\end{lemma}
\begin{proof}
See Appendix~\ref{A6}.
\end{proof}

Lemma~\ref{L6-2} shows that for any mapping distribution $\mathcal{D} \ne \mathcal{D}^*$, we can still guarantee that the margin difference between generating any two contention thresholds $W$ and $\hat{W}$ given the prompt $P$ under the pre-trained mapping distribution $\mathcal{D}$ is at least half of the margin difference under the ground-truth mapping distribution $\mathcal{D}^*$. In other words, the LLM is still able to distinguish the correct contention window threshold even if we change the input distribution by concatenating examples.

Based on Lemma~\ref{L6-2}, we are now ready to prove our main ICL result. Denote $\Delta_{\mathcal{D^*}}$ as the minimal difference between Bayes Optimal Classifier prediction and any mapping $f \in \mathcal{D^*}$. We have the following.

\begin{theorem}\label{T6.3}
    Given the margin $\Delta_{\mathcal{D^*}}$ of the prompt mapping distribution satisfying $\Delta_{\mathcal{D^*}} > 1 - \alpha^2$ and denote $c:= 1 - \sqrt{\frac{1-\alpha^2}{\Delta_{\mathcal{D^*}}}}$. If in-context example number $M$ satisfies
   \begin{align*}
     M \geq  \max\bigg\{\frac{(\ln \frac{1}{q}) (16\ell^2)(\ln^2\beta)}{KL^2(\mathbb{P}_{f}, \mathbb{P}_{f^*})}, \frac{-2\ln(\frac{\epsilon}{2(1-\frac{c}{2})^{-1}\alpha^{-2}\beta^{-T}\gamma^{-1}})}{\min_{f} KL(\mathbb{P}_{f}, \mathbb{P}_{f^*})+8\ln(\alpha \beta)}\bigg\}, 
   \end{align*} 
   and the ICL prediction loss $\mathcal{L}$ in \eqref{eq9} satisfies
    \begin{align*}
      Pr(\mathcal{L} - \text{BER} \leq \epsilon) \geq 1-q,
    \end{align*}
   where $\text{BER}=\mathbb{E}_{x\sim\mathcal{D}^*}[1-\max_{W}\mathbb{P}_{\mathcal{D}^*}(W|x)]$ stands for the lowest possible error rate of the Bayes optimal classifier given the ground-truth mapping. Besides, the throughput loss $\Delta U$ due to an erroneous prompt is given as follows:
    \begin{align*}
      Pr\bigg(\Delta U \leq \mathcal{O}\bigg(\frac{T_P\bar{N}}{8T_\sigma}\epsilon^\frac{1}{2} \bar{W}\bigg)\bigg) \geq 1-q.
    \end{align*}
\end{theorem}
\begin{proof}
See Appendix~\ref{A7}.
\end{proof}
Theorem~\ref{T6.3} implies that a small error $\epsilon$ on ICL prediction requires the example number $M$ large enough.
Since the contention window threshold can be erroneous in each collision case in any prompt, the upper bound in Theorem~\ref{T6.3} is further constrained by the maximum contention window threshold $\bar{W}$ compared with Theorem~\ref{prop1} in the perfect prompt case.

\section{Simulation Experiments}\label{S7}

In Section~\ref{S7.1}, we introduce our experimental settings of the co-used channel. In Section~\ref{S7.2}, we introduce our experimental results
in the training stage of four steps as illustrated in Figure~\ref{fig4-1}. In Sections~\ref{S7.3} and \ref{S7.4}, we introduce our experimental results in the testing stage with unknown and heavy node densities.

\subsection{Experiment Settings}\label{S7.1}

\begin{table}
\caption{Network settings (unit of $\mu s$) for experiments in Section~\ref{S7}. We fix the channel bit rate as 1Mbps. }
\begin{center}
\begin{tabular}{|c|c|c|c|c|c|c|c|c|}
\hline
$T_\sigma$& $T_{\text{DIFS}}$& $T_{\text{SIFS}}$ & $T_\delta$ & $T_{\text{ACK}}$ & $T_{\text{Header}}$ & $T_{\text{P}}$ & $T_s$ & $T_c$ \\
\hline
50& 128& 28 & 1  & 240 & 400 & 8184 & 8982 & 8783\\
\hline
\end{tabular}
\label{tab1}
\end{center}
\end{table}

Following the CSMA literature (e.g., \cite{bianchi2000performance}, \cite{zheng2014performance}), we consider the same settings as the DCF in 802.11 protocols. We summarize the network environment settings in Table~\ref{tab1}. Note that the settings are essential for obtaining the closed-form throughput $U$ in \eqref{eq6} for ICL data collection and further comparisons. 

To simulate the dynamic environment of node density, we use prompts generated under low-density environments to train the transformer in Section~\ref{S7.2}. We then use the trained transformer to test under unknown node density environments of even hundreds of nodes in Sections~\ref{S7.3} and \ref{S7.4}.

\subsection{Experimental Results in the Training Stage}\label{S7.2}

We consider node density environments $N$$\in$$\{2, 3, 4, 5, 6\}$ for ICL data collection. As in Step I of Figure~\ref{fig4-1}, for each node density $N$, the AP system offline collects $M=9$ data points $\{(x_k, W_k^*)\}_{k=0}^8$ to construct the prompt, where $x_k = (k, T_P, T_{s}, T_{c})$, $k \in [K]$ denotes the current collision number and the maximum collision number is set as $K=8$. Note that the node density $N$ is not shown in any $x_k$.  To obtain the contention window thresholds $\{W_k^*\}_{k=0}^8$ in a prompt with a node density $N$, the system first maximizes the throughput $U$ in \eqref{eq6} for obtaining the optimal packet transmission probability $\tau^*$ (e.g., using the branch-and-bound algorithm). Then, it uses the $\tau^*$ to solve $\{W_k^*\}_{k=0}^8$ according to \eqref{eq1} by applying MILP algorithms like golden section search.

As in Step II of Figure~\ref{fig4-1}, we construct $S=5$ prompts corresponding to 5 different node densities $N$$\in$$\{2, 3, 4, 5, 6\}$ for training the transformer parameters $\theta$, where each prompt contains 8 example points $(x_k, W_k^*)$ and a query collision case $x_q^s$. These prompts are then embedded as in Step III of Figure~\ref{fig4-1}. Regarding Step IV of Figure~\ref{fig4-1} for transformer training, we set the step size of Algorithm 1 as $\eta = 0.05$. We consider $x_q^s=x_k$ to be the same for all 5 prompts and obtain our ICL prediction $\hat{W}_k=\hat{W}_q^s$ in \eqref{eqW} of each contention window threshold $W_k^*$, $k \in [K]$. Then, wireless devices download the collision-threshold pair table to use online for CSMA, fully compatible with legacy devices.

\begin{figure}
    \centering
    \includegraphics[scale=0.25]{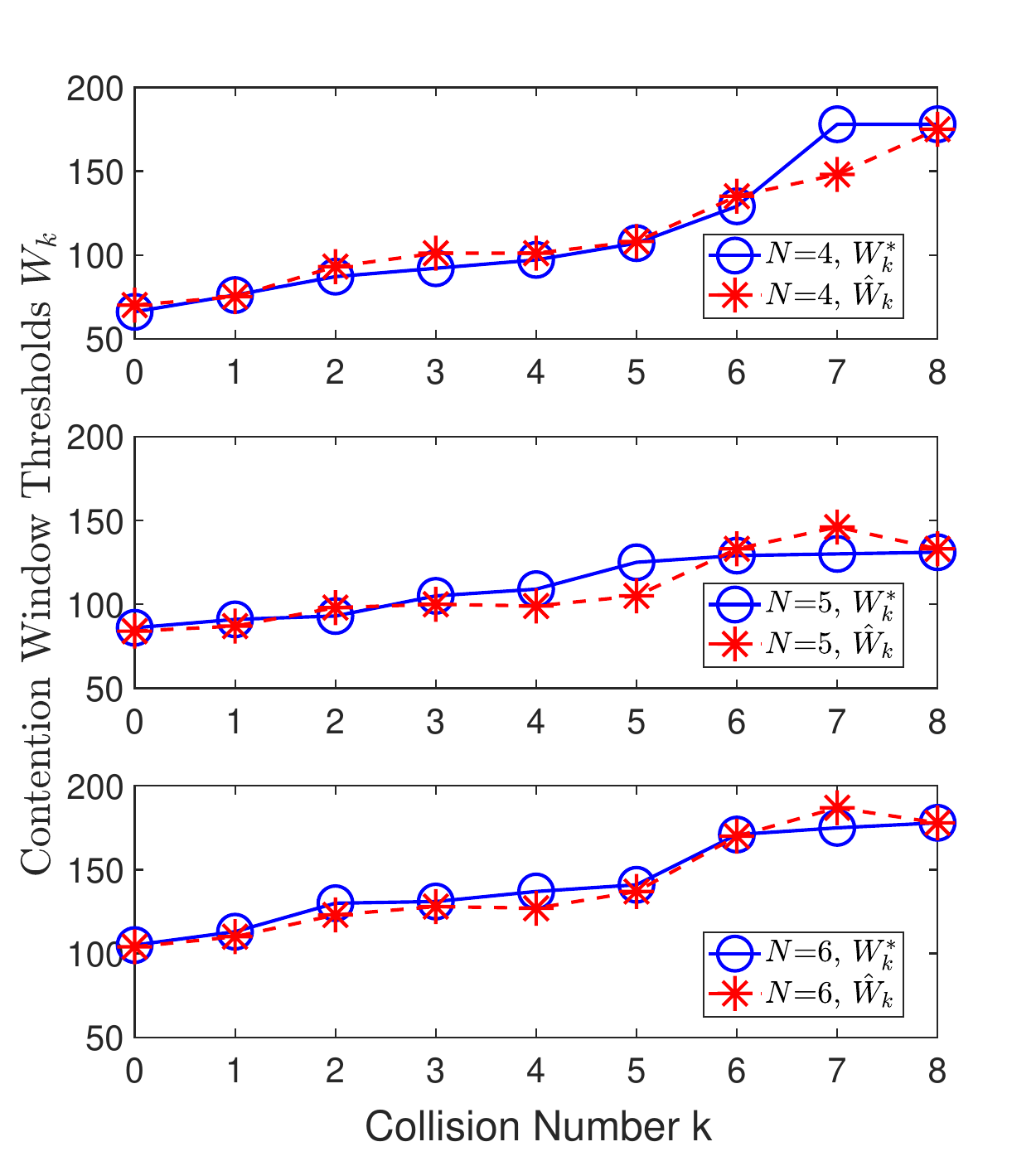}
    \caption{Contention window thresholds $\{W_k^*\}_{k=0}^K$ at the optimum and $\{\hat{W}_k\}_{k=0}^K$ of our approach versus the collision number $k$, respectively. Here we fix the maximum collision number $K=8$ and change the node density $N \in \{4, 5, 6\}$ across the three subfigures as included in the training stage.}
    \label{fig7-0}
\end{figure}
\begin{figure}
    \centering
    \includegraphics[scale=0.3]{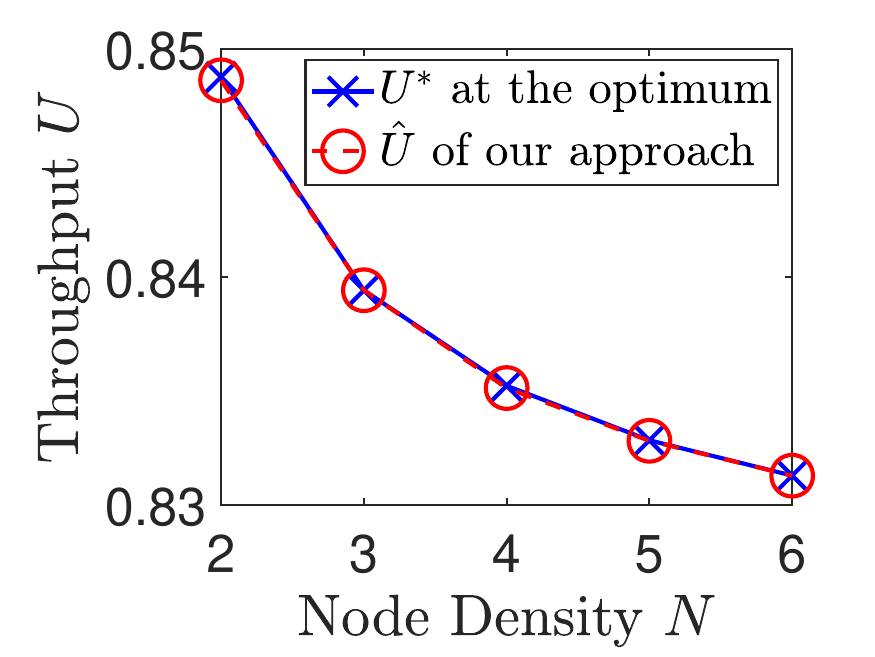}
    \caption{The throughput $U^*$ at the optimum and $\hat{U}$ of our transformer-based ICL approach versus the node density $N$, respectively. Here we check for each node density $N \in \{2, 3, 4, 5, 6\}$ as included in the training stage.}
    \label{fig7-01}
\end{figure}

Figure~\ref{fig7-0} plots contention window thresholds $\{W_k^*\}_{k=0}^K$ at the optimum and $\{\hat{W}_k\}_{k=0}^K$ of our transformer-based ICL approach versus the collision number $k$, respectively. It shows that our approach can approximate each optimal contention window threshold in the node density environment of the training stage, which is consistent with Theorem~\ref{T5.3} in Section~\ref{S5}.


Figure~\ref{fig7-01} plots the throughput $U^*$ at the optimum and $\hat{U}$ of our transformer-based ICL approach versus the node density $N$, respectively. It indicates that our approach can nearly achieve the optimal throughput in each node density $N$ considered in the training stage, which is consistent with the limited throughput loss in  Proposition~\ref{prop1} of Section~\ref{S5}. 


\subsection{Experimental Results in the Testing Stage}\label{S7.3}

In the following, we want to test the adaptiveness of our transformer-based ICL approach to dynamic node density environments. We consider a challenging case where each contention threshold $\tilde{W}_k$ is erroneous in each testing prompt as in Section~\ref{S6}. In particular, we define that a prompt is with $b\%$ error if each erroneous threshold $\tilde{W}_k$ is randomly realized from the set of $\{(100-b)W_k^*, (100+b)W_k^*\}$ regarding the ground-truth $W_k^*$, where $b \in (0, 100)$. We follow the same steps as in the training stage to construct prompts for testing.

\begin{figure}
    \centering
    \includegraphics[scale=0.35]{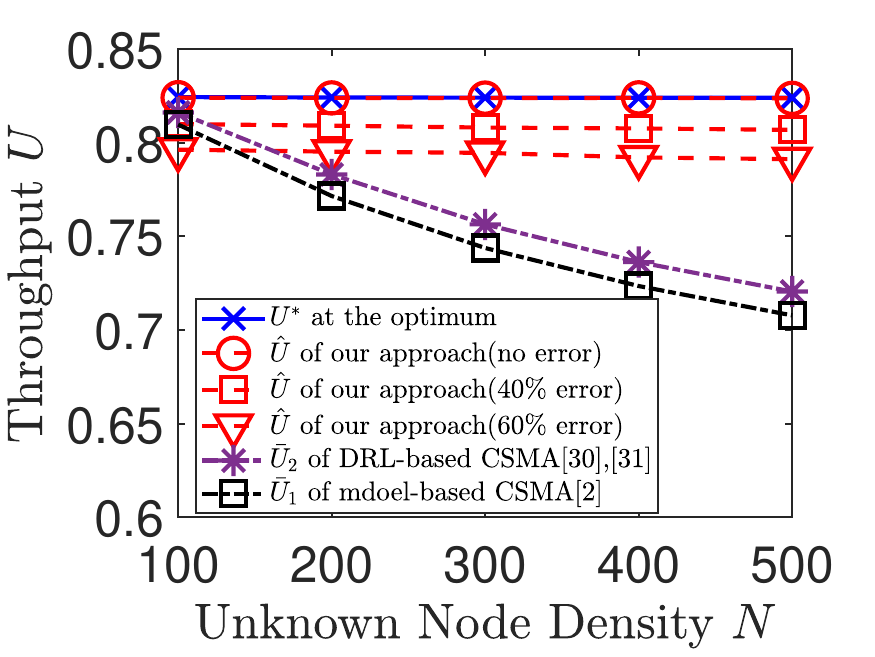}
    \caption{The throughput $U^*$ at the optimum, $\hat{U}$ of our transformer-based ICL approach with no error, 40\% error and 60\% error, $\bar{U}_{1}$ of the model-based benchmark, and $\bar{U}_{2}$ of the DRL-based benchmark versus the node density $N$, respectively. Here we optimize benchmarks under an approximated node density of $\hat{N} = 50$ and change the unknown $N \in [100, 500]$.}
    \label{fig7-2}
\end{figure}

Figure~\ref{fig7-2} plots the throughput $U^*$ at the optimum, $\hat{U}$ of our transformer-based ICL approach with no error, 40\% error and 60\% error, $\bar{U}_{2}$ of the DRL-based benchmark, and $\bar{U}_{1}$ of the model-based benchmark versus the node density $N$ even scales up to hundreds, respectively. It shows that our approach with no error still achieves the near-optimal throughput even under high and unknown node densities, which is consistent with Theorem~\ref{T5.3} of Section~\ref{S5}. Further, Figure~\ref{fig7-2} indicates that the throughput loss of our approach with erroneous prompts are still limited even under high and unknown node densities, which is consistent with Theorem~\ref{T6.3} of Section~\ref{S6}.

Figure~\ref{fig7-2} also shows that our transformer-based ICL approach with no error always outperforms the benchmark schemes and our transformer-based ICL approach with large errors still outperforms either as long as the unknown node density $N$ is larger than 200. The throughput difference between benchmarks and the optimum enlarges as the gap between estimated node density and the ground truth increases, which is consistent with Theorem~\ref{L4} of Section~\ref{S3}.

\begin{figure}
    \centering
    \includegraphics[scale=0.35]{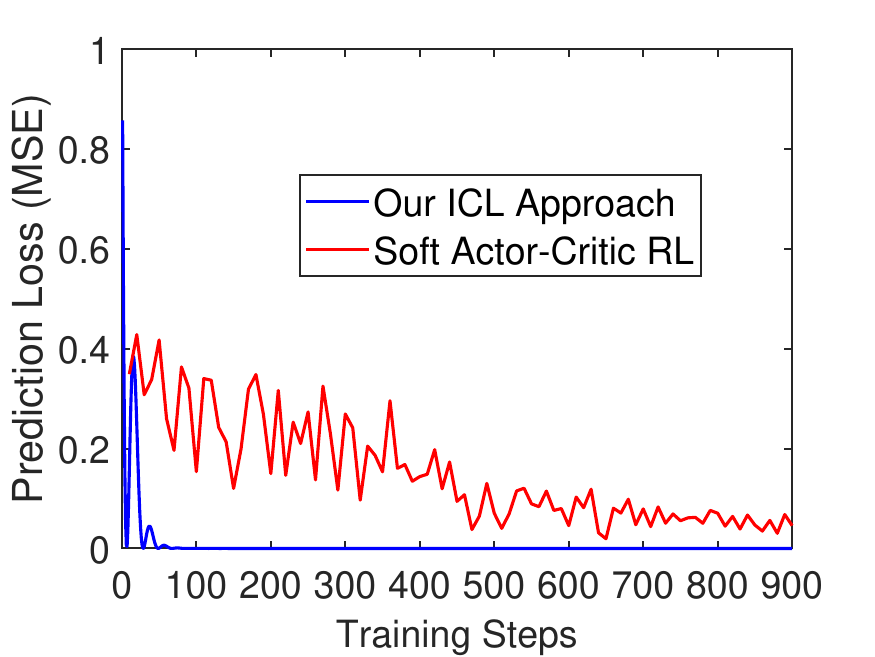}
    \caption{Prediction loss versus training steps for our ICL approach and soft actor-critic RL, respectively.}
    \label{fig-cg}
\end{figure}
\begin{figure}
    \centering
    \includegraphics[scale=0.3]{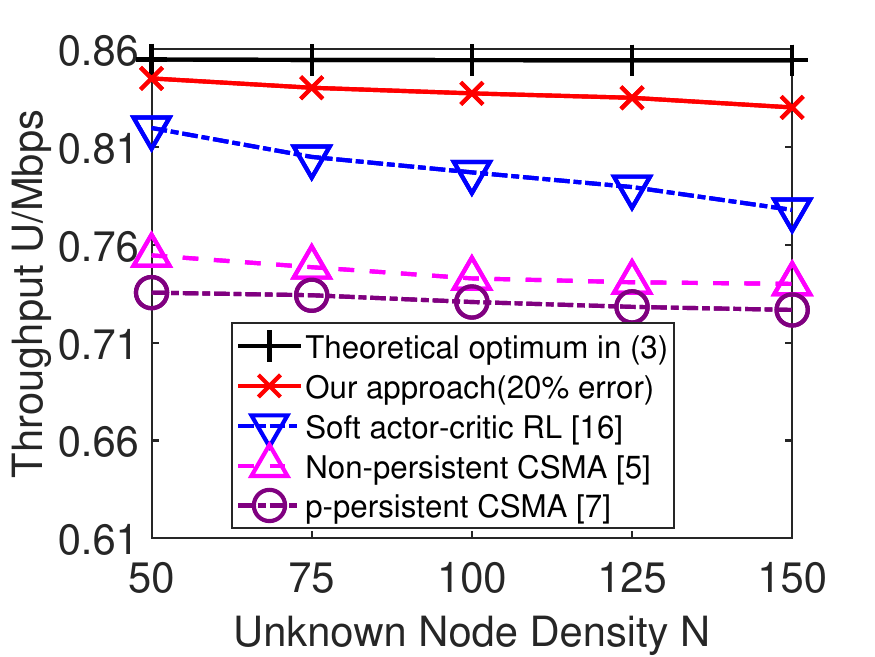}
    \caption{Throughput versus unknown node density $N$ under different approaches in NS-3.38. Wireless devices are randomly placed on a 20m‐radius circle around an AP and send UDP packets every 1 ms to a server on the AP.}
    \label{fig-ns}
\end{figure}

\subsection{More Practical Experiments on NS-3}\label{S7.4}

For further comparison with recent CSMA advances, we now use NS-3.38 to resemble a real-world experiment. Wireless devices are randomly placed on a 20m-radius circle around a WiFi AP. The 802.11b link runs at 1 Mbps with 16 dBm transmit power and a 7 dB noise. Each device sends 1029-byte UDP packets every 1 ms to a server on the AP via port 8000. We use NS-3’s FlowMonitor to record the throughput as the total successfully received application-layer payload divided by the effective transmission time. We train our transformer only under small-scale node densities $N\in[2,6]$ with 20\% erroneous prompts, optimize model-based approaches under an estimated node density of $N$=10, and optimize DRL-based approaches to adapt to unknown $N\in[50,150]$. We implement the Soft Actor–Critic (SAC) algorithm with both actor and twin-critic networks as two-layer MLPs (128 units per layer, ReLU activations), outputting a tanh-squashed Gaussian policy (log-std clamped to $[-20, 2]$). The critic and the actor are trained with Adam at a learning rate of $1 \times 10^{-4}$, using a discount factor $\gamma = 0.99$, soft-target updates ($\tau = 0.005$), and automatic entropy tuning (initial $\alpha \approx 0.1$, targeting $-\text{dim}(\mathbf{a})$). We employ a small replay buffer (capacity $2000$) to prioritize recent transitions, with full-batch updates every 20 steps followed by buffer clearance. Our ICL approach incurs the same learning rate.

Figure~\ref{fig-cg} shows prediction loss versus training steps for our ICL approach and soft actor-critic RL (the remaining DRL-based approaches are too slow to be comparable with ours), respectively. Figure~\ref{fig-cg} shows that the soft actor-critic RL incurs at least 650 training steps to achieve a stable convergence with $<$0.1 loss. In contrast, our approach attains zero loss within the first 100 steps, indicating a much faster convergence.

Figure~\ref{fig-ns} shows throughput versus unknown node density $N$ under different approaches in NS-3.38. It indicates our ICL approach's near-optimal throughput performance within 3\% loss, greatly outperforming recent advances in the CSMA literature. Though SAC's throughput is the closest to ours, Figure~\ref{fig-cg} has indicated our approach's much faster convergence for ease of implementation.





\section{Conclusion}\label{S8}

In this paper, we study LLM transformer-based ICL for optimizing channel access. We propose a transformer-based ICL optimizer to pre-collect collision-threshold data examples and a query collision case. They are constructed as a prompt as the input for the transformer to learn the pattern, which then generates a predicted CWT. To train the transformer for effective ICL, we develop an efficient algorithm and guarantee a near-optimal CWT prediction within limited training steps, ensuring a converged throughput loss. As it may be hard to gather perfect data examples in practice, we further extend to allow erroneous examples in the prompt. We prove that our optimizer still incurs limited prediction and throughput losses from the optimum. Experimental results on NS-3 further demonstrate our approach's fast convergence and near-optimal throughput over existing model-based and DRL-based approaches.  










\appendix
\onecolumn

\section{Proof of Theorem~\ref{L4}}\label{A1}

Before proving the theorem, we introduce a useful lemma, which is obtained by checking the first derivative of throughput $U$ in \eqref{eq6}.
\begin{lemma}\label{A1-L1}
    The optimal solution $\tau^*$ to maximize $U(\tau)$ in \eqref{eq6} satisfies $\tau^* < 1/N$.
\end{lemma}

To prove the lower bound in Theorem~\ref{L4}, we first prove that there exists a $C_1 > 0$ such that $|\tau(N)-\tau(\hat{N})| \geq C_1 |N -\hat{N}|$. Then, we prove that there exists a $C_2 > 0$ such that $|U(\tau(N)) - U(\tau(\hat{N}))| \geq C_2 |\tau(N)-\tau(\hat{N})|$.

To find the exact $C_1$, we define
\begin{align*}
    D(N,\tau) := (1-(1-\tau)^{N-1})\sum_{k=0}^{K-1}(1-(1-\tau)^{N-1})^k W_k + (1-(1-\tau)^{N-1})^K W_K + 1.
\end{align*}
According to \eqref{eq1}, we have $\tau = \frac{2}{D}$.  Differentiating $\tau = \frac{2}{D}$ with respect to $N$, we have
\begin{align*}
    \frac{\partial \tau}{\partial N} = -\frac{\tau\left(\frac{\partial D}{\partial N} + \frac{\partial D}{\partial \tau}\frac{\partial \tau}{\partial N}\right)}{D}.
\end{align*}
After rewriting the above equality, we obtain that 
\begin{align*}
    \bigg|\frac{\partial \tau}{\partial N}\bigg| = \frac{|\tau\frac{\partial D}{\partial N}|}{|D + \tau\frac{\partial D}{\partial \tau}|}.
\end{align*}
According to the Mean Value Theorem, there exists a $\tau_0 \in (0, 1)$ such that
\begin{align*}
    |\tau(N)-\tau(\hat{N})| = \bigg|\frac{\partial \tau}{\partial N}\big|_{\tau=\tau_0}\bigg| |N -\hat{N}|.
\end{align*}
Therefore, to prove that $|\tau(N)-\tau(\hat{N})| \geq C_1 |N -\hat{N}|$, it is enough to prove that $|\frac{\partial \tau}{\partial N}| \geq C_1$. This motivates us to find a lower bound of the numerator of $|\frac{\partial \tau}{\partial N}|$ and an upper bound of the denominator of $|\frac{\partial \tau}{\partial N}|$. The partial derivative of $D(N,\tau)$ on $N$ is:
\begin{align}\label{A1-eq1}
    \frac{\partial D}{\partial N} = \frac{\partial D}{\partial p}\frac{\partial p}{\partial N}= \left(\sum_{k=0}^{K-1}(k+1)p^k W_k + Kp^{K-1}W_K\right) \bigg(-(1-\tau)^{N-1}\ln(1-\tau)\bigg) 
    \geq (1-\tau)^{N-1}\tau,
\end{align}
where the inequality holds due to 
\begin{align*}
    \sum_{k=0}^{K-1}(k+1)p^k W_k + Kp^{K-1}W_K \geq \sum_{k=0}^{K-1}(k+1)p^k W_k \geq \sum_{k=0}^{K-1}(k+1)p^k  \geq 1
\end{align*}
and $-\ln(1-\tau) \geq \tau$ for $\tau \in (0,1)$. The partial derivative of $D(N,\tau)$ on $\tau$ is:
\begin{align}\label{A1-eq2}
    \frac{\partial D}{\partial \tau} = \frac{\partial D}{\partial p}\frac{\partial p}{\partial \tau} = \left(\sum_{k=0}^{K-1}(k+1)p^k W_k + Kp^{K-1}W_K\right)(N-1)(1-\tau)^{N-2} \leq \bar{W}(N-1)(1-\tau)^{N-2}\left(\frac{K(K+1)}{2} + K\right)
\end{align}
due to $W_k \leq \bar{W}$ and $p \leq 1$ for $k \in [K]$. Based on \eqref{A1-eq1} and \eqref{A1-eq2}, we obtain a lower bound of $|\frac{\partial \tau}{\partial N}|$ as follows:
\begin{align}\label{A1-eq3}
    \left|\frac{\partial \tau}{\partial N}\right| \geq \frac{\tau (1-\tau)^{N-1}\tau}{\frac{2}{\tau} + \tau\bar{W}(N-1)(1-\tau)^{N-2}\left(\frac{K(K+1)}{2} + K\right)}.
\end{align}
Since $W_k \leq \bar{W}$, we have $\tau = \frac{2}{D} \geq \frac{2}{\bar{W}+1}$. Together with $\tau \leq \frac{1}{N} \leq \frac{1}{2}$, $(1-\tau)^{N-1} \geq e^{-1}$ for $\tau \leq \frac{1}{N}$ and $(1-\tau)^{N-2} \leq 1$, we further bound $\left|\frac{\partial \tau}{\partial N}\right|$ according to \eqref{A1-eq3} as follows:
\begin{align*}
    \left|\frac{\partial \tau}{\partial N}\right| \geq \frac{e^{-1} \frac{4}{(\bar{W}+1)^2}}{ \frac{4}{\bar{W}+1} + \frac{1}{2} \bar{W} (\bar{N}-1)  \left(\frac{K(K+1)}{2} + K\right) } = \Theta \bigg(\frac{1}{\bar{N}\bar{W}^3K^2}\bigg) = C_1.
\end{align*}

To find the exact $C_2$, similar to the analysis of $\tau(n)$, we can check that $U(\tau)$ in \eqref{eq6} satisfies $|U'(\tau)| \geq \frac{T_\sigma}{T_c^2}$.
According to \eqref{eq6}, we can check that $U'(\tau)$ is bounded at $\tau=0$ and $\tau=1$, which implies that $U'(\tau)$ is bounded for $\tau \in [0, 1]$, leading to 
\begin{align}\label{A1-eq4}
    |U(\tau(n)) - U(\tau(\hat{n}))| = |U'(\tau_0)| \cdot |\tau(n) - \tau(\hat{n})| 
    \geq \frac{T_\sigma}{T_c^2} |\tau(n) - \tau(\hat{n})|, 
\end{align}
implying $C_2 = \frac{T_\sigma}{T_c^2}$. Using the explicit $C_1$ and $C_2$, we finally have
\begin{align*}
    |U(\tau(n)) - U(\tau(\hat{n}))| \geq \frac{T_\sigma}{T_c^2} |\tau(n) - \tau(\hat{n})| \geq \Theta \bigg(\frac{T_\sigma}{\bar{N}K^2\bar{W}^3T_c^2}\bigg) |N - \hat{N}|.
\end{align*}

\section{Proof of Lemma~\ref{lemma:loss}}\label{A0}
We rewrite the prediction error function $\mathcal{L}^{(t)}(\theta)$ in (\ref{eq7}) into:
\begin{align*}
    \mathcal{L}^{(t)}(\theta)=&\frac{1}{2}\sum_{k=1}^K\mathbb{E}\left[\mathbf{1}\{x_q^s=x_k\}(\hat{W}^s_q-f^s(x_k))^2\right]\\
    =&\frac{1}{2}\sum_{k=1}^K\mathbb{E}\left[\mathbf{1}\{x_q^s=x_k\}\Big(\sum_{n\in[K]}\text{Attn}_n^{(t)}f^s(x_n)-f^s(x_k)\Big)^2\right]\\
    =&\frac{1}{2}\sum_{k=1}^K\mathbb{E}\left[\mathbf{1}\{x_q^s=x_k\}\Big(\sum_{n\neq k}\text{Attn}_n^{(t)}\big(f^s(x_n)-f^s(x_k)\big)\Big)^2\right]\\
    = &\frac{1}{2}\sum_{k=1}^K\mathbb{E}\left[\mathbf{1}\{x_q^s=x_k\}\Big(1-\text{Attn}_k^{(t)}\Big)^2\Theta(L^2\Delta^2)\right],
\end{align*}
where the last equality is because of $\sum_{n\neq k}\text{Attn}_n^{(t)}=1-\text{Attn}_k^{(t)}$ and $\big|f^s(x_n)-f^s(x_k)\big|=\Theta(L\Delta)$. 
 
\section{Proof of Proposition~\ref{prop_convergence}}\label{A2}

Recall that there is a probability of $\Theta(\frac{1}{K})$ that each token $x_i$ is a noisy version of feature $x_k\in\mathcal{X}$. Let $P^{(t)}_{1:N}$ denote the collection of input tokens for $P^{(t)}$, i.e., $\{x_i\}_{i=1}^N$. We derive the following lemma to characterize the size of the token set $\mathcal{X}^s_k$ for $P^{(t)}_{1:N}$. 
\begin{lemma}\label{lemma:balanced_Vk}
Suppose $K^3=\mathcal{O}(N)$. For some constant $c\geq \sqrt{\frac{20K^3}{N}}$, define 
\begin{align}
    \mathcal{E}^*:=\left\{P^{(t)}_{1:N}:|\mathcal{X}^s_k|\in\left[p_kN-\frac{cN}{K},p_kN+\frac{cN}{K}\right]\text{ for }k\in[K]\right\}.
\end{align}
Then, we have
\begin{align*}
    \mathbb{P}(P_{1:N}^{(t)}\in\mathcal{E}^*)\geq 1-3\exp{\left(-\frac{c^2N}{25K^2}\right)}.
\end{align*}
\end{lemma}
\begin{proof}
Note that $|\mathcal{X}_k|\sim\text{multinomial}(N,p_1,\cdots, p_K)$. Let $\delta=\frac{c}{K}$, such that $\frac{\delta^2}{20}\geq \frac{K}{N}$. 

According to tail bound of the multinomial distribution, we obtain
\begin{align*}
    \mathbb{P}\left(\sum_{i=1}^K\big||\mathcal{X}^s_k|-\mathbb{E}(|\mathcal{X}^s_k|)\big|>c\frac{N}{K}\right)\leq 3\exp(-\frac{c^2N}{25K^2}).
\end{align*}
Since $\mathbb{E}[|\mathcal{X}_k|]=p_kN$, we have
\begin{align*}
    \mathbb{P}\left(\cap_{i=1}^K\left\{\big||\mathcal{X}_k|-\mathbb{E}(|\mathcal{X}_k|)\big|>c\frac{N}{K}\right\}\right)\leq \mathbb{P}\left(\sum_{i=1}^K\big||\mathcal{X}_k|-\mathbb{E}(|\mathcal{X}_k|)\big|>c\frac{N}{K}\right)\leq 3\exp(-\frac{c^2N}{25K^2}),
\end{align*}
which completes the proof of Lemma~\ref{lemma:balanced_Vk}.
\end{proof}
For ease of exposition, we define $u_k:=(p_k-\delta_0)K$ and $U_k:=(p_k+\delta_0)K$, which satisfy $u_k=\Theta(1)$ and $U_k=\Theta(1)$, given $p_k=\Theta(\frac{1}{K})$ and $\delta_0=\mathcal{O}(\frac{1}{K})$. Then for any $P_{1:N}^{(t)}$ belonging to $\mathcal{E}^*$, we have $|\mathcal{X}_k|\in[\frac{u_kN}{K},\frac{UK}{N}]=\Theta(\frac{N}{K})$ with a probability close to $1$. We then prove Proposition~\ref{prop_convergence} based on Lemma~\ref{lemma:balanced_Vk}.

Given $k,k'\in[K]$ with $k'\neq k$, for $t\geq 0$ we define the bilinear attention weights as follows:
\begin{align*}
    A_k:=x_k^\top Q^{(t)} x_k, \quad \alpha_k^{(t)}=-x_k^{\top}\nabla_{Q^{(t)}} \mathcal{L}(Q^{(t)})x_k,\\
    B_{k,k'}:=x_n^\top Q^{(t)} x_k,\quad \beta_{k,k'}^{(t)}=-x_n^{\top}\nabla_{Q^{(t)}} \mathcal{L}(Q^{(t)})x_k.
\end{align*}
Note that the bilinear attention weights characterize the attention scores between any two tokens in eq. (\ref{SA}). Then our Algorithm~1 achieves convergence if and only if the bilinear attention weights converge, i.e., the gradients $\alpha_k^{(t)}=\mathcal{O}(\delta_0)$ and $\beta_{k,k'}^{(t)}=\mathcal{O}(\delta_0)$. Consequently, we next prove that the two conditions are equivalent with $1-\text{Attn}_k^{(t)}=\mathcal{O}(\delta_0)$.

In the following, we first derive the expressions of $\alpha_k^{(t)}$ and $\beta_{k,k'}^{(t)}$, respectively. By gradient descent (GD) update, we have
\begin{align*}
    A_k^{(t+1)}:=A_k^{(t)}+\eta \alpha_k^{(t)}\\
    B_{k,k'}^{(t+1)}:=B_{k,k'}^{(t)}+\eta \beta_{k,k'}^{(t)}.
\end{align*}
Next, we try to derive $\alpha_k^{(t)}$ and $\beta_{k,k'}^{(t)}$. We calculate
\begin{align*}
    x_{k'}^\top \nabla_{Q^{(t)}} \mathcal{L} x_k = \mathbb{E}\Big[\mathbf{1}\{x_q^s=x_k\}(\hat{W}_q^s-f^s(x_k))\sum_{i,j\in[N]}\text{attn}^{(t)}_i\text{attn}_j^{(t)}f^s(x_i)x_{k'}^{\top}(E_i^x-E_j^x)\Big],
\end{align*}
which is because ${E_{N+1}^x}^\top x_k\neq 0$ if and only if $x_q^s=x_k$. Then we continue calculating
\begin{align*}
    x_{k'}^\top \nabla_{Q^{(t)}} \mathcal{L} x_k =&\mathbb{E}\Big[\mathbf{1}\{x_q^s=x_k\}(\hat{W}_q^s-f^s(x_k))\sum_{m,n\in[K]}\sum_{i\in\mathcal{X}_m}\sum_{j\in\mathcal{X}_n}\text{attn}_i^{(t)}\text{attn}_j^{(t)}f^s(x_i)x_{k'}^\top (x_m-x_n)\Big]\\
    =&\mathbb{E}\Big[\mathbf{1}\{x_q^s=x_k\}(\hat{W}_q^s-f^s(x_k))\sum_{n\in[K]}\sum_{i\in\mathcal{X}_{k'}}\sum_{j\in\mathcal{X}_n}\text{attn}^{(t)}_i\text{attn}^{(t)}_jf^s(x_i)x_{k'}^\top (x_{k'}-x_n)\Big]\\&+\mathbb{E}\Big[\mathbf{1}\{x_q^s=x_k\}(\hat{W}_q^s-f^s(x_k))\sum_{m\in[K]}\sum_{i\in\mathcal{X}_m}\sum_{j\in\mathcal{X}_{k'}}\text{attn}^{(t)}_i\text{attn}^{(t)}_jf^s(x_i)x_{k'}^\top (x_m-x_{k'})\Big]\\
    =&\mathbb{E}\Big[\mathbf{1}\{x_q^s=x_k\}(\hat{W}_q^s-f^s(x_k))\text{Attn}^{(t)}_{k'} f^s(x_{k'})\sum_{n\in[K]}\text{Attn}^{(t)}_{n}\Big]\\ 
    &-\mathbb{E}\Big[\mathbf{1}\{x_q^s=x_k\}(\hat{W}_q^s-f^s(x_k))\text{Attn}^{(t)}_{k'} \sum_{m\in[K]}\text{Attn}^{(t)}_{m}f^s(x_{m})\Big]\\
    =&\mathbb{E}\Big[\mathbf{1}\{x_q^s=x_k\}(\hat{W}_q^s-f^s(x_k))\text{Attn}^{(t)}_{k'} \sum_{m\in[K]}\text{Attn}^{(t)}_{m}(f^s(x_{k'})-f^s(x_m))\Big].
\end{align*}

Since $\hat{W}^s_q=\sum_{i\in[N]}\text{attn}_if^s(x_i)=\sum_{m\in[K]}\text{Attn}^{(t)}_m f^s(x_m)$, we obtain
\begin{align*}
    x_{k'}^\top \nabla_{Q^{(t)}} \mathcal{L} x_k &=-\mathbb{E}\Big[\mathbf{1}\{x_q^s=x_k\}\text{Attn}^{(t)}_{k'} \sum_{n\in[K]}\sum_{m\in[K]}\text{Attn}^{(t)}_{m}\text{Attn}^{(t)}_{n}(f^s(x_{k})-f^s(x_n))(f^s(x_{k'})-f^s(x_{m}))\Big]\\
    &=-\mathbb{E}\Big[\mathbf{1}\{x_q^s=x_k\}\text{Attn}^{(t)}_{k'} (f^s(x_k)-\sum_{n\in[K]}\text{Attn}^{(t)}_{n}f^s(x_n))(f^s(x_{k'})-\sum_{m\in[K]}\text{Attn}^{(t)}_m f^s(x_m))\Big].
\end{align*}

If $k'=k$, we obtain
\begin{align}
    \alpha_k^{(t)}&=-x_{k}^\top \nabla_{Q^{(t)}} \mathcal{L} x_k\notag\\
    &=\mathbb{E}\Big[\mathbf{1}\{x_q^s=x_k\}\text{Attn}_{k}^{(t)}\Big(f^s(x_k)-\sum_{n\in[K]}\text{Attn}_n^{(t)}f^s(x_n)\Big)^2\Big].\label{alpha_k(t)}
\end{align}
As $\alpha_k^{(t)}\geq 0$ is always true, $A_k^{(t)}$ increases with $t$.

If $k'\neq k$, we obtain
\begin{align}
    \beta_{k,k'}^{(t)}&=-x_{k'}^\top \nabla_{Q^{(t)}} \mathcal{L} x_k\notag\\
    &=\mathbb{E}\Big[\mathbf{1}\{x_q^s=x_k\}\text{Attn}_{k'}^{(t)}\Big(f^s(x_k)f^s(x_{k'})-(f^s(x_k)+f^s(x_k'))\sum_{m\in[K]}\text{Attn}_m^{(t)} f^s(x_m)+\big(\sum_{m\in[K]}\text{Attn}_m^{(t)} f^s(x_m)\big)^2\Big)\Big]\notag\\
    &=\mathbb{E}\Big[\mathbf{1}\{x_q^s=x_k\}\text{Attn}_{k'}^{(t)} (f^s(x_k)-\sum_{n\in[K]}\text{Attn}_{n}^{(t)}f^s(x_n))(f^s(x_{k'})-\sum_{m\in[K]}\text{Attn}_m^{(t)} f^s(x_m))\Big].\label{beta_k(t)}
\end{align}

Based on the expressions of $\alpha_k^{(t)}$ and $\beta_{k,k'}^{(t)}$ above, we next prove the system convergences if and only if $1-\text{Attn}_k^{(t)}=\mathcal{O}(\delta_0)$. We first prove that if $1-\text{Attn}_k^{(t)}=\mathcal{O}(\delta_0)$, the system achieves convergence. Then we prove that if the system achieves convergence, the attention score satisfies $1-\text{Attn}_k^{(t)}=\mathcal{O}(\delta_0)$ for any $k\in[K]$. Then based on the two conclusions, we can derive Proposition~\ref{prop_convergence}.

We suppose $x_q^s=x_k$ in the following. If $|\text{Attn}_k^{(t)}-1|=\mathcal{O}(\delta_0)$, we have $\sum_{n\in[K],n\neq k}\text{Attn}^{(t)}_n=1-\text{Attn}_k^{(t)}=\mathcal{O}(\delta_0)$, which means $\text{Attn}_{k'}=\mathcal{O}(\delta_0)$ for any $k'\neq k$. Then based on (\ref{alpha_k(t)}) and (\ref{beta_k(t)}), we calculate
\begin{align*}
    \alpha_k^{(t)}=&\mathbb{E}\Big[\text{Attn}_k^{(t)}\Big(f^s(x_k)-\sum_{n\in[K]}\text{Attn}_n^{(t)}f^s(x_n)\Big)^2\Big]\\
    =&\mathbb{E}\Big[\Theta(1)\cdot\Big(\sum_{n\in[K]}\big(\text{Attn}_n^{(t)}f^s(x_k)-\text{Attn}_n^{(t)}f^s(x_n)\big)\Big)^2\Big]\\
    =&\mathcal{O}(\delta_0).
\end{align*}
Similarly, we calculate
\begin{align*}
    \beta_{k,k'}^{(t)}=&\mathbb{E}\Big[\text{Attn}_{k'}^{(t)} (f^s(x_k)-\sum_{n\in[K]}\text{Attn}_{n}^{(t)}f^s(x_n))(f^s(x_{k'})-\sum_{m\in[K]}\text{Attn}_m^{(t)} f^s(x_m))\Big]\\
    =&\mathbb{E}\Big[\text{Attn}_{k'}^{(t)} (f^s(x_k)-\text{Attn}_k f^s(x_k)+\mathcal{O}(\delta_0))(f^s(x_{k'})-\sum_{m\in[K]}\text{Attn}_m^{(t)} f^s(x_m))\Big]\\
    =&\mathcal{O}(\delta_0).
\end{align*}
Consequently, the system has achieved the convergence state.

If the system has achieved the convergence state at time $T_s$, we prove (\ref{convergence}) by contradiction. At time $t>T_s$, we assume that the attention score $\text{Attn}_k^{(t)}$ of the query token $x_q^s=x_k$ satisfies $1-\text{Attn}_k^{(t)}=\Theta(1)$. Then there must exist another $k'\in[K]$ with $k'\neq k$ so that $\text{Attn}_{k'}^{(t)}=\Theta(1)$ for the current prompt. Then we calculate
\begin{align*}
    \alpha_k^{(t)}=&\mathbb{E}\Big[\text{Attn}_k^{(t)}\Big(f^s(x_k)-\sum_{n\in[K]}\text{Attn}_n^{(t)}f^s(x_n)\Big)^2\Big]\\
    =&\mathbb{E}\Big[\Theta(1)\cdot\Big(\sum_{n\in[K]}\big(\text{Attn}_n^{(t)}f^s(x_k)-\text{Attn}_n^{(t)}f^s(x_n)\big)\Big)^2\Big]\\
    \geq &\mathbb{E}\Big[\Theta(1)\cdot\big(\text{Attn}_{k'}^{(t)}(f^s(x_k)-f^s(x_{k'}))\big)^2\Big]\\
    =&\Omega(\delta_0),
\end{align*}
where the second equality is because of $\text{Attn}_k^{(t)}=\Theta(1)$, the last equality is because of $\|x_k-x_{k'}\|=\Theta(\Delta)$ and Assumption~1. Given the gradient $\alpha_k^{(t)}=\Omega(\delta_0)$, we have $|\text{Attn}_{k}^{(t+1)}-\text{Attn}_{k}^{(t)}|=\Omega(\Delta^2)$, which is contradicted with the convergence state $\alpha_k^{(t)}=\mathcal{O}(\delta_0)$. Consequently, if the system has achieved the convergence state, (\ref{convergence}) always holds, which completes the proof of Proposition~\ref{prop_convergence}.

\section{Proof of Theorem~\ref{T5.3}}\label{A3}
In this proof, we suppose $x_q^s=x_k$. We first prove that the attention score $\text{Attn}_k^{(t)}$ in (\ref{W_output}) increases to $\text{Attn}_k^{(T^*)}=\Omega(\frac{1}{1+\delta_0\epsilon})$ at time $T^*=\Theta(\frac{K\log(K\epsilon^{-1})}{\eta\delta_0^2 L^2\Delta^2})$. Then we prove at time $T^*$, this achieved attention score ensures that the prediction loss satisfies $\mathcal{L}(\theta^{T^*})=\mathcal{O}(\epsilon^2)$.

\subsection{Growth of the target attention score $\text{Attn}_k^{(t)}$}\label{proof_T5.3A}
We study the growth of $\text{Attn}_k^{(t)}$ in two learning stages:
\begin{itemize}
    \item In the first stage, where $t\in\{1,\cdots, T_1\}$ with $T_1=\Theta(\frac{K\log(K)}{\eta L^2\Delta^2})$, the bilinear attention weight $A_k^{(t)}=x_k^\top Q^{(t)}x_k$ increases at a rate of $\Theta(\frac{\eta L^2\Delta^2}{K})$. After the end of the first stage, we have  $\text{Attn}_k^{(T_1+1)}=\Omega(\frac{1}{1+\delta_0})$.
    \item In the second stage, where $t\in\{T_1+1,\cdots, T^*\}$ with $T^*=\Theta(\frac{K\log(K\epsilon^{-1})}{\eta\delta_0^2 L^2\Delta^2})$, $A_k^{(t)}$ increases at a rate of $\Theta(\frac{\eta \delta_0^2L^2\Delta^2}{K})$, and we obtain $\text{Attn}_k^{(T^*)}=\Omega(\frac{1}{1+\delta_0\epsilon})$ at the end of the second stage.
\end{itemize}

We first rewrite the gradient $\alpha_k^{(t)}$ as follows:
\begin{align*}
    \alpha_k^{(t)}=&\mathbb{E}\Big[\mathbf{1}\{x_q^s=x_k\}\text{Attn}_{k}^{(t)}\Big(f^s(x_k)-\sum_{n\in[K]}\text{Attn}_n^{(t)}f^s(x_n)\Big)^2\Big]\\
    =&\mathbb{E}\Big[\mathbf{1}\{x_q^s=x_k\}\text{Attn}_{k}^{(t)}\Big(\sum_{n\in[K]}\text{Attn}_n^{(t)}\big(f^s(x_k)-f^s(x_n)\big)\Big)^2\Big]\\
    \geq &\mathbb{E}\Big[\mathbf{1}\{x_q^s=x_k\}\text{Attn}_{k}^{(t)}\Big(\sum_{n\in[K]}\text{Attn}_n^{(t)}\min_{k,n\in[K]}|f^s(x_k)-f^s(x_n)|\Big)^2\Big]\\
    =&\mathbb{E}\Big[\mathbf{1}\{x_q^s=x_k\}\text{Attn}_{k}^{(t)}\Big(1-\text{Attn}_k^{(t)}\Big)^2\Theta(L^2\Delta^2)\Big],
\end{align*}
where the last equality is due to Assumption 1.
Consequently, we obtain $\alpha_k^{(t)}=\mathbb{E}\Big[\mathbf{1}\{x_q^s=x_k\}\text{Attn}_{k}^{(t)}\Big(1-\text{Attn}_k^{(t)}\Big)^2 \Theta(L^2\Delta^{2})\Big]$. 

For the first stage with $t\in\{1,\cdots, T_1\}$, based on the gradient expression above, we calculate
\begin{align*}
    \alpha_k^{(t)}=&\mathbb{E}\Big[\mathbf{1}\{x_q^s=x_k\}\text{Attn}_{k}^{(t)}\Big(1-\text{Attn}_k^{(t)}\Big)^2\Theta(L^2\Delta^2)\Big]\\
    = &p_k\cdot  \mathbb{E}\left[\text{Attn}_{k}^{(t)}\Big(1-\text{Attn}_k^{(t)}\Big)^2\Big| x_q^s=x_k\}\right]\cdot \Theta(L^2\Delta^2)\\
    =&\Theta(\frac{L^2\Delta^2}{K}).
\end{align*}
where the last equality is because of $p_k=\Theta(\frac{1}{K}),\text{Attn}_{k}^{(t)}= \Theta(1)$ and $1-\text{Attn}_{k}^{(t)}=\Theta(1)$ for $t\in\{1,\cdots, T_1\}$. 
For any $k\neq n$, we calculate
\begin{align*}
    |\beta_{k,n}^{(t)}|&=\mathbb{E}\Big[\mathbf{1}\{x_q^s=x_k\}\text{Attn}_{n}^{(t)} |f^s(x_k)-\sum_{j\in[K]}\text{Attn}_{j}^{(t)}f^s(x_j)|\cdot |f^s(x_{n})-\sum_{m\in[K]}\text{Attn}_m^{(t)} f^s(x_m)|\Big]\\
    &=p_k\cdot \mathbb{E}\Big[\text{Attn}_{n}^{(t)} \Big|\sum_{j\in[K]}(\text{Attn}_{j}^{(t)}f^s(x_k)-\text{Attn}_{j}^{(t)}f^s(x_j))\Big|\cdot \Big|\sum_{m\in[K]}(\text{Attn}_m^{(t)}f^s(x_{n})-\text{Attn}_m^{(t)} f^s(x_m))\Big|\Big]\\
    &<p_k\cdot \mathbb{E}\Big[\text{Attn}_{n}^{(t)} \cdot\sum_{j\in[K]}\text{Attn}_{j}^{(t)}|f^s(x_k)-f^s(x_j)|\cdot \sum_{m\in[K]}\text{Attn}_m^{(t)}|f^s(x_{n})-f^s(x_m))|\Big]\\
    &=p_k\cdot \mathbb{E}\left[\text{Attn}_{n}^{(t)} \cdot (1-\text{Attn}_{k}^{(t)} )\cdot (1-\text{Attn}_{n}^{(t)} )\cdot \Theta(L^2\Delta^2)\right]\\
    &=\mathcal{O}(\frac{L^2\Delta^2}{K^2}),
\end{align*}
where the inequality is derived by union bound, and the last equality is because of $|f^s(x_{k})-f^s(x_j)|=|f^s(x_{n})-f^s(x_m)|=\Theta(L\Delta)$ for any $j\neq k$ and $m\neq n$, respectively, and the last equality is because of $\text{Attn}_{n}^{(t)}=\Theta(\frac{1}{K})$, $p_k=\Theta(\frac{1}{K})$, and $1-\text{Attn}_{k}^{(t)}<1$.

Then we calculate
\begin{align*}
    A_k^{(T_1+1)}&=A_k^{(T_1)}+\eta \alpha_k^{(T_1)}=A_k^{(T_1-1)}+\eta \alpha_k^{(T_1-1)}+\eta \alpha_k^{(T_1)}\\
    &=\cdots=A_k^{(0)}+\eta\cdot \Theta(\frac{L^2\Delta^2}{K})\cdot T_1=\Theta(\log(K)).
\end{align*}
Given $|\beta_{k,m}^{(t)}|=\mathcal{O}(\frac{L^2C^2}{K^2})$, we can similarly calculate 
\begin{align*}
    B_{k,m}^{(T_1+1)}&=B_{k,m}^{(T_1)}+\eta \beta_{k,m}^{(T_1)}\\
    &\leq |B_{k,m}^{(T_1)}|+\eta |\beta_{k,m}^{(T_1)}|\\
    &\leq |B_{k,m}^{(0)}|+\eta \cdot \mathcal{O}(\frac{L^2\Delta^2}{K^2})\cdot T_1\\
    &=\mathcal{O}(\frac{\log(K)}{K}).
\end{align*}
Consequently, we have $B_{k,m}^{(T_1+1)}=\mathcal{O}(\frac{\log(K)}{K})$.
Finally, we calculate the attention score
\begin{align*}
    \text{Attn}_k^{(t)}&=\frac{|\mathcal{X}_k|e^{x_k^\top Q^{(t)} x_k}}{\sum_{j\in[N]}e^{{E_j^x}^\top Q^{(t)}x_k}}\\
    &=\frac{|\mathcal{X}_k|e^{x_k^\top Q^{(t)} x_k}}{\sum_{m\neq k}|\mathcal{X}_m|e^{{x_m}^\top Q^{(t)}x_k}+|\mathcal{X}_k|e^{x_k^\top Q^{(t)} x_k}}\\
    &=\frac{1}{\sum_{m\neq k}\frac{|\mathcal{X}_m|}{|\mathcal{X}_k|}\exp{(B_{k,m}^{(t)}-A_k^{(t)})}+1}.\\
    &\geq \frac{1}{\mathcal{O}(\frac{1}{K})(\frac{N}{|\mathcal{X}_k|}-1)+1}\\&\geq \frac{1}{\mathcal{O}(\frac{1}{u_k}-\frac{1}{K})+1}\\&=\Omega(\frac{1}{1+\delta_0}),
\end{align*}
where the first inequality is because of $\exp{(B_{k,m}^{(t)}-A_k^{(t)})}\leq \exp{(\frac{\log(K)}{K}-\log(K))}\leq \mathcal{O}(\frac{1}{K})$, and the last equality is because of $\frac{1}{u_k}-\frac{1}{K}=\Theta(\delta_0)$ derived in Lemma~\ref{lemma:balanced_Vk}.

For the second stage with $t\in\{T_1+1,\cdots, T^*\}$, we can use the similar way to calculate
\begin{align*}
    \alpha_k^{(t)}=\Theta(\frac{\delta^2 L^2\Delta^2}{K}), \quad |\beta_{k,m}^{(t)}|=\mathcal{O}(\frac{\delta^2L^2\Delta^2}{K^2}).
\end{align*}
Then at the end of the second stage, we obtain
\begin{align*}
    A_k^{(T^*)}=\Theta(\log(K\epsilon^{-1})), \quad B_{k,m}^{(T^*)}=\Theta(\frac{\log(K\epsilon^{-1})}{K}).
\end{align*}
We further calculate 
\begin{align*}
    \text{Attn}_k^{(T^*)}\geq \frac{1}{O(\frac{\epsilon}{K})(\frac{N}{|\mathcal{X}_k|}-1)+1}\geq \frac{1}{O(\epsilon)\cdot O(\frac{1}{u_k}-\frac{1}{K})+1}=\Omega(\frac{1}{1+\epsilon\delta_0}),
\end{align*}

\subsection{Convergence of prediction loss $\mathcal{L}(\theta)$}

Suppose $x_q^s=x_k$ at time $T^*$. Based on the conclusion $1-\text{Attn}_k^{(T^*)}=\mathcal{O}(\epsilon\delta_0)$ derived in Appendix~\ref{proof_T5.3A} and $\mathcal{L}^{(t)}(\theta)$ in Lemma~\ref{lemma:loss}, we finally calculate
\begin{align*}
    \mathcal{L}^{(T^*)}(\theta)&=\frac{1}{2}\sum_{k=1}^K\mathbb{E}\left[\mathbf{1}\{x_q^s=x_k\}\Big(1-\text{Attn}_k^{(T^*)}\Big)^2\Theta(L^2\Delta^2)\right]\\
    &=\mathbb{E}\left[\Big(1-\text{Attn}_k^{(T^*)}\Big)^2\Theta(L^2\Delta^2)\right]\\
    &=\mathcal{O}(\epsilon^2),
\end{align*}
where the last equality is because of $\Theta(L^2\Delta^2) \cdot \mathcal{O}(\epsilon^2\delta_0^2)=\mathcal{O}(\epsilon^2)$ given $L\leq \Theta(\frac{1}{L\Delta})$.
This completes the proof of Theorem~\ref{T5.3}.

\section{Proof of Lemma~\ref{L5.4}}\label{A4}

To prove the upper bound in Lemma~\ref{L5.4}, we first prove that there exists a $C_3 > 0$ such that $|\tau(W_k)-\tau(\hat{W}_k)| \leq C_3 |W_k -\hat{W}_k|$. Then, we prove that there exists a $C_4 > 0$ such that $|U(\tau(W_k)) - U(\tau(\hat{W}_k))| \leq C_4 |\tau(W_k)-\tau(\hat{W}_k)|$.

To find the exact $C_3$, we define:
\begin{align*}
    f^s(\tau, \boldsymbol{W}) := \tau(1-\tau)^{N-1}\sum_{k=0}^{K-1}(1-(1-\tau)^{N-1})^k W_k + \tau (1-(1-\tau)^{N-1})^KW_K + \tau - 2,
\end{align*}
where $\boldsymbol{W} =\{W_k\}_{k=0}^K$ is the vector of contention window thresholds. We can check that 
\begin{align*}
    \bigg|\frac{\partial f^s(\tau, \boldsymbol{W})}{\partial \tau}\bigg| \geq \bigg|\frac{\partial f^s(\tau, \boldsymbol{W})}{\partial \tau}\bigg|_{W_k=1}\bigg| \geq 2. 
\end{align*}
The derivative of $F$ on $W_k$ is 
\begin{align*}
    \frac{\partial f^s(\tau, \boldsymbol{W})}{\partial W_k} = 
    \begin{cases}  
    \tau(1-\tau)^{N-1}(1-(1-\tau)^{N-1})^k, &\text{if} \ k= 0, \cdots, K-1, \\
    \tau(1-(1-\tau)^{N-1})^K, &\text{if} \ k=K.
    \end{cases}
\end{align*}
Note that
\begin{align}\label{A4-eq1}
    \tau(1-\tau)^{N-1}(1-(1-\tau)^{N-1})^k \leq \tau(1-\tau)^{N-1} \leq \frac{1}{N}\bigg(1-\frac{1}{N}\bigg)^{N-1} \leq \frac{1}{4},
\end{align}
where the first inequality holds due to $(1-(1-\tau)^{N-1})^k \leq 1$, the second due to $\tau(1-\tau)^{N-1}$ increasing with $\tau \leq \frac{1}{N}$ (this bound $1/N$ comes from Lemma~\ref{A1-L1}),
and the last due to $\frac{1}{N}(1-\frac{1}{N})^{N-1}$ decreasing with $N \geq 2$. Further, we have
\begin{align}\label{A4-eq2}
    \tau(1-(1-\tau)^{N-1})^K \leq \frac{1}{N}\bigg(1-\bigg(1-\frac{1}{N}\bigg)^{N-1}\bigg)^K \leq \frac{1}{2}\bigg(1-\bigg(1-\frac{1}{\bar{N}}\bigg)^{\bar{N}-1}\bigg)^K \leq \frac{1}{2} \bigg(1-\frac{1}{e}\bigg)^K < \frac{1}{3},
\end{align}
where the first inequality holds due to $\tau(1-(1-\tau)^{N-1})^K$ increasing with $\tau \leq \frac{1}{N}$, the second due to $N \in [2, \bar{N}]$ and $(1-(1-\frac{1}{N})^{N-1})^K$ increasing with $N$, the third due to $(1-\frac{1}{\bar{N}})^{\bar{N}-1} \geq 1/e$, the last due to $(1-\frac{1}{e})^K < \frac{2}{3}$. According to \eqref{A4-eq1} and \eqref{A4-eq2}, we can now upper bound $\frac{\partial f^s(\tau, \boldsymbol{W})}{\partial W_k}$ as follows:
\begin{align*}
    \frac{\partial f^s(\tau, \boldsymbol{W})}{\partial W_k} \leq \frac{1}{4}.
\end{align*}
Based on the Implicit Function Theorem, we have
\begin{align*}
    \bigg|\frac{\partial \tau}{\partial W_k}\bigg| = \frac{|\frac{\partial f^s(\tau, \boldsymbol{W})}{\partial W_k}|}{|\frac{\partial f^s(\tau, \boldsymbol{W})}{\partial \tau}|} \leq \frac{1}{8},
\end{align*}
implying
\begin{align*}
    |\tau(W_k)-\tau(\hat{W}_k)| \leq \frac{1}{8} |W_k - \hat{W}_k|
\end{align*}
according to the Mean Value Theorem. Then we can set $C_3 = \frac{1}{8}$. Following a similar analysis, we can upper bound $|U'(\tau)|$ as 
\begin{align*}
|U'(\tau)| \leq \frac{T_P\bar{N}}{T_\sigma} = C_4,
\end{align*}
implying
\begin{align*}
  |U(\tau(W_k))-U(\tau(\hat{W}_k))| 
  \leq \frac{T_P\bar{N}}{T_\sigma} |\tau(W_k) - \tau(\hat{W}_k)|
  \leq \frac{T_P\bar{N}}{8T_\sigma}  |W_k - \hat{W}_k|.
\end{align*}

\section{Proof of Theorem~\ref{prop1}}\label{A5}


According to Theorem~\ref{T5.3}, we have
\begin{align}\label{A5-eq1}
    \mathcal{L}(\theta^{(T^*)}) = \mathbb{E} \bigg[ \big(\hat{W}_q^s - W_q^s \big)^2\bigg] \leq \mathcal{O}(\epsilon^2).
\end{align}
Thus, we have
\begin{align*}
  \Delta U &= \mathbb{E}\bigg[ \bigg(U({W}_q^s) - U(\hat{W}_q^s)\bigg)\bigg] \\
  &= \mathbb{E}|U({W}_q^s) - U(\hat{W}_q^s)| \\
 &\leq  \mathbb{E} \bigg[\frac{T_P\bar{N}}{8T_\sigma} \cdot | 
 {W}_q^s - \hat{W}_q^s|\bigg] \\
 &= \frac{T_P\bar{N}}{8T_\sigma}  \mathbb{E}    
 \big|{W}_q^s - \hat{W}_{q}^s\big| \\
 &\leq \frac{T_P\bar{N}}{8T_\sigma} \bigg(   \mathbb{E} 
 \big({W}_q^s - \hat{W}_{q}^s\big)^2\bigg)^{\frac{1}{2}} \\
 &\leq \frac{T_P\bar{N}}{8T_\sigma}   \bigg(   \mathbb{E} 
 \big({W}_q^s - \hat{W}_{q}^s\big)^2\bigg)^{\frac{1}{2}} \\
&\leq \frac{T_P\bar{N}}{8T_\sigma} \cdot 
 \mathcal{O}(\epsilon) = \mathcal{O}\bigg(\frac{T_P\bar{N}\epsilon}{8T_\sigma}\bigg),
\end{align*}
where the first inequality holds due to Lemma~\ref{L5.4}, the second and the third hold due to Jensen's inequality, and the last due to \eqref{A5-eq1}. We then finish the proof.

\section{Proof of Lemma~\ref{L6-2}}\label{A6}

Before the formal proof, let us introduce a useful lemma in the following.
\begin{lemma}\label{A6-L1}
Suppose that the minimum KL-divergence between our ICL mapping $f$ and the ground-truth $f^*$ satisfies $\min_{f} KL(\mathbb{P}_{f}, \mathbb{P}_{f^*})$$> -8\ln(\alpha \beta)$. If the number of in-context data examples $M$ is long enough as 
\begin{align*}
    M \geq \max\bigg\{\frac{-(\ln q) (16\ell^2)(\ln^2\beta)}{KL^2(\mathbb{P}_{f}, \mathbb{P}_{f^*})}, \frac{-2\ln\mu}{\min_{f} KL(\mathbb{P}_{f}, \mathbb{P}_{f^*})+8\ln(\alpha \beta)}\bigg\}
\end{align*}
for any $\mu > 0$, $q \in (0, 1)$  and any mapping $f \ne f^*$, we have
\begin{align*}
    Pr\bigg(\frac{\mathbb{P}_{f}(P)}{\mathbb{P}_{f^*}(P)} < \mu\bigg) \geq 1 - q.
\end{align*}
\end{lemma}

In Section~\ref{A6-1}, we first prove Lemma~\ref{A6-L1}. Then, we prove Lemma~\ref{L6-2} in Section~\ref{A6-2}.

\subsection{Proof of Lemma~\ref{A6-L1}}\label{A6-1}

Given
\begin{align*}
    \alpha \mathbb{P}_{f}(s_2|s_1 \oplus ``\mathrm{\backslash n}") \leq  \mathbb{P}_{f}(s_2) 
    \leq \frac{1}{\alpha} \mathbb{P}_{f}(s_2|s_1 \oplus ``\mathrm{\backslash n}"), \ \alpha \in (0, 1].
\end{align*}
with $s_1 = x_1 \oplus W_1 \oplus \cdots \oplus x_m \oplus W_m$ and $s_2 = x_{m+1} \oplus W_{m+1}$, $m \leq M-1$, we have
\begin{align*}
    \alpha \leq \frac{\mathbb{P}_{f}(s_1 \oplus ``\mathrm{\backslash n}") \cdot \mathbb{P}_{f}(s_2)}{ \mathbb{P}_{f}(s_1 \oplus ``\mathrm{\backslash n}" \oplus s_2)} \leq \frac{1}{\alpha}.
\end{align*}
By multiplying $\frac{\mathbb{P}_{f}(s_1 \oplus ``\mathrm{\backslash n}") \cdot \mathbb{P}_{f}(s_2)}{ \mathbb{P}_{f}(s_1 \oplus ``\mathrm{\backslash n}" \oplus s_2)}$ for all the possible $s_1$ and $s_2$, we obtain the following inequality:
\begin{align}\label{A6-eq1}
    \alpha^M \leq \frac{\prod_{m=1}^M \mathbb{P}_{f}(x_m \oplus W_m \oplus ``\mathrm{\backslash n}")}{\mathbb{P}_{f}(x_1 \oplus W_1 \oplus ``\mathrm{\backslash n}" \oplus \cdots \oplus x_M \oplus W_M \oplus ``\mathrm{\backslash n}")} \leq \alpha^{-M}.
\end{align}

Further, we have
\begin{align}\label{A6-eq2}
    \mathbb{P}_{f}(x_m \oplus W_m) = \mathbb{P}_{f}(x_m) \mathbb{P}_{f}(W_m|x_m) > \mathbb{P}_{f}(x_m) \mathbb{P}_{f}(W_m^*|x_m) \mathbb{P}_{f}(W_m|W_m^*) =  \mathbb{P}_{f}(x_m \oplus W_m^*) \mathbb{P}_{f}(W_m|W_m^*) > \mathbb{P}_{f}(x_m \oplus W_m^*) \cdot \beta,
\end{align}
where the inequality holds due to the assumption in Section~\ref{S6.1}. According to \eqref{A6-eq2}, we also have 
\begin{align}\label{A6-eq3}
    \mathbb{P}_{f}(x_m \oplus W_m^*) > \mathbb{P}_{f}(x_m \oplus W_m) \cdot \beta.
\end{align}
Based on \eqref{A6-eq2} and \eqref{A6-eq3}, we have
\begin{align}\label{A6-eq4}
    \beta < \frac{\mathbb{P}_{f}(x_m \oplus W_m^*)}{\mathbb{P}_{f}(x_m \oplus W_m)} < \beta^{-1}.
\end{align}

Since
\begin{align}\label{A6-eq5}
    \mathbb{P}_{f}(x_m \oplus W_m) \geq \mathbb{P}_{f}(x_m \oplus W_m \oplus ``\mathrm{\backslash n}") = \mathbb{P}_{f}(x_m \oplus W_m) \cdot \mathbb{P}_{f}(``\mathrm{\backslash n}"|x_m \oplus W_m) > \mathbb{P}_{f}(x_m \oplus W_m) \cdot \beta,
\end{align}
where the inequality holds due to the assumption in Section~\ref{S6.1}. Further,
\begin{align}\label{A6-eq6}
    \mathbb{P}_{f}(x_m \oplus W_m \oplus ``\mathrm{\backslash n}") < \mathbb{P}_{f}(x_m \oplus W_m) < \mathbb{P}_{f}(x_m \oplus W_m) \cdot \beta^{-1}
\end{align}
due to $\beta \in (0, 1)$. According to \eqref{A6-eq5} and \eqref{A6-eq6}, we have
\begin{align}\label{A6-eq7}
    \beta < \frac{\mathbb{P}_{f}(x_m \oplus W_m)}{\mathbb{P}_{f}(x_m \oplus W_m \oplus ``\mathrm{\backslash n}")} < \beta^{-1}.
\end{align}

According to \eqref{A6-eq1}, \eqref{A6-eq4} and \eqref{A6-eq7}, we can now obtain that
\begin{align}\label{A6-eq8}
    \frac{\prod_{m=1}^M \mathbb{P}_{f}(x_m \oplus W_m^*)}{\mathbb{P}_{f}(x_1 \oplus W_1 \oplus ``\mathrm{\backslash n}" \oplus \cdots \oplus x_M \oplus W_M \oplus ``\mathrm{\backslash n}")} &= \frac{\prod_{m=1}^M \mathbb{P}_{f}(x_m \oplus W_m \oplus ``\mathrm{\backslash n}")}{\mathbb{P}_{f}(x_1 \oplus W_1 \oplus ``\mathrm{\backslash n}" \oplus \cdots \oplus x_M \oplus W_M \oplus ``\mathrm{\backslash n}")} \cdot \frac{\prod_{m=1}^M \mathbb{P}_{f}(x_m \oplus W_m)}{\prod_{m=1}^M \mathbb{P}_{f}(x_m \oplus W_m \oplus ``\mathrm{\backslash n}")} \cdot \frac{\prod_{m=1}^M \mathbb{P}_{f}(x_m \oplus W_m^*)}{\prod_{m=1}^M \mathbb{P}_{f}(x_m \oplus W_m)} \nonumber \\
    &\in [\alpha^M \beta^{2M}, \alpha^{-M}\beta^{-2M}].
\end{align}

Denote $P = x_1 \oplus W_1 \oplus ``\mathrm{\backslash n}" \oplus \cdots \oplus x_M \oplus W_M \oplus ``\mathrm{\backslash n}"$. We have
\begin{align*}
    \ln \frac{\mathbb{P}_{f}(P)}{\mathbb{P}_{f^*}(P)} &= \ln \frac{\mathbb{P}_{f}(P)}{\prod_{m=1}^M \mathbb{P}_{f}(x_m \oplus W_m^*)} + \ln \frac{\prod_{m=1}^M \mathbb{P}_{f}(x_m \oplus W_m^*)}{\prod_{m=1}^M \mathbb{P}_{f^*}(x_m \oplus W_m^*)} + \ln \frac{\prod_{m=1}^M \mathbb{P}_{f^*}(x_m \oplus W_m^*)}{\mathbb{P}_{f^*}(P)} \\
    &\leq \ln \alpha^{-M}\beta^{-2M} + \ln \frac{\prod_{m=1}^M \mathbb{P}_{f}(x_m \oplus W_m^*)}{\prod_{m=1}^M \mathbb{P}_{f^*}(x_m \oplus W_m^*)} + \ln \alpha^{-M}\beta^{-2M} \\
    &= 4M \ln \alpha^{-1}\beta^{-1} + \sum_{m=1}^M \ln \frac{\mathbb{P}_{f}(x_m \oplus W_m^*)}{\mathbb{P}_{f^*}(x_m \oplus W_m^*)},
\end{align*}
where the inequality holds due to \eqref{A6-eq8}. Note that
\begin{align}\label{A6-eq9}
    \mathbb{E}\bigg[\frac{1}{M}\sum_{m=1}^M\ln\frac{\mathbb{P}_{f}(x_m \oplus W_m^*)}{\mathbb{P}_{f^*}(x_m \oplus W_m^*)} \bigg] = -KL(\mathbb{P}_{f^*}, \mathbb{P}_{f}),
\end{align}
and
\begin{align}\label{A6-eq10}
    \bigg| \ln \frac{\mathbb{P}_{f}(x_m \oplus W_m^*)}{\mathbb{P}_{f^*}(x_m \oplus W_m^*)} \bigg| = \bigg| \sum_{l=1}^\ell \ln \frac{\mathbb{P}_{f}(x_m^{l+1}|x_m^{1:l})}{\mathbb{P}_{f^*}(x_m^{l+1}|x_m^{1:l})} \bigg| \leq \sum_{l=1}^\ell \bigg| \ln \frac{\mathbb{P}_{f}(x_m^{l+1}|x_m^{1:l})}{\mathbb{P}_{f^*}(x_m^{l+1}|x_m^{1:l})} \bigg| \leq \sum_{l=1}^\ell \ln \beta^{-1} = \ell \ln \beta^{-1},
\end{align}
where the second inequality holds due to 
\begin{align*}
    \frac{\mathbb{P}_{f}(x_m^{l+1}|x_m^{1:l})}{\mathbb{P}_{f^*}(x_m^{l+1}|x_m^{1:l})} \leq \frac{1}{\mathbb{P}_{f^*}(x_m^{l+1}|x_m^{1:l})} \leq \frac{1}{\beta} \ \text{with} \ \mathbb{P}_{f^*}(x_m^{l+1}|x_m^{1:l}) > \beta.
\end{align*}
Based on \eqref{A6-eq9} and \eqref{A6-eq10}, according to the Hoeffding Inequality, we have
\begin{align}\label{A6-eq11}
    Pr\bigg( \frac{\mathbb{P}_{f}(P)}{\mathbb{P}_{f^*}(P)} \leq e^{-M(KL(\mathbb{P}_{f^*}, \mathbb{P}_{f}) - \mu' - 4\ln \alpha^{-1} \beta^{-1})} \bigg) \geq 1 - e^{\frac{-2M \mu'}{(2\ell \ln \beta)^2}}.
\end{align}
We take $\mu' = \frac{1}{2} KL(\mathbb{P}_{f^*}, \mathbb{P}_{f})$. Therefore, for any $(\mu, q)$ satisfies 
\begin{align}\label{A6-eq12}
    \mu > e^{\frac{M}{2} (KL(\mathbb{P}_{f^*}, \mathbb{P}_{f}) - 8\ln \alpha^{-1} \beta^{-1}) }, \ q > e^{-\frac{M (KL(\mathbb{P}_{f^*}, \mathbb{P}_{f}))^2}{8 \ell^2 \ln^2 \beta}},
\end{align}
we always have 
\begin{align*}
    Pr\bigg( \frac{\mathbb{P}_{f}(P)}{\mathbb{P}_{f^*}(P)} \leq \mu) \geq 1 - q.
\end{align*}
After rewriting \eqref{A6-eq12}, we obtain the condition on $M$ as follows:
\begin{align*}
    M \geq \max\bigg\{\frac{-(\ln q) (16\ell^2)(\ln^2\beta)}{KL^2(\mathbb{P}_{f}, \mathbb{P}_{f^*})}, \frac{-2\ln\mu}{\min_{\phi} KL(\mathbb{P}_{f}, \mathbb{P}_{f^*})+8\ln(\alpha \beta)}\bigg\}.
\end{align*}

\subsection{Proof of Lemma~\ref{L6-2}}\label{A6-2}
Denote $P' = x_1 \oplus W_1 \oplus ``\mathrm{\backslash n}" \oplus \cdots \oplus x_M \oplus W_M \oplus ``\mathrm{\backslash n}"$ and $P = P' \oplus x$. We have
\begin{align}\label{A6-eq13}
    \mathbb{P}_{\mathcal{D}}(W|P' \oplus x) - \mathbb{P}_{\mathcal{D}}(W^*|P' \oplus x) = \frac{\sum_{f \in \mathcal{D}} Pr(f|\mathcal{D}) (\mathbb{P}_{f}(P' \oplus x \oplus W) -  \mathbb{P}_{f}(P' \oplus x \oplus W^*)) }{ \sum_{f \in \mathcal{D}} Pr(f|\mathcal{D}) \mathbb{P}_{f}(P' \oplus x)  }.
\end{align}
According to the assumption in Section~\ref{S6.1}, we have
\begin{align}
    &\alpha \leq \frac{\mathbb{P}_{f}(P' \oplus x \oplus W)}{\mathbb{P}_{f}(P')\mathbb{P}_{f}(x \oplus W)} \leq \alpha^{-1}, \nonumber \\
    &\mathbb{P}_{f}(P' \oplus x) \leq \alpha^{-1} \mathbb{P}_{f}(P' ) \mathbb{P}_{f}(x), \label{A6-eq14}
\end{align}
which implies
\begin{align}
    &\mathbb{P}_{f}(P' \oplus x \oplus W) \geq \alpha \mathbb{P}_{f}(P') \mathbb{P}_{f}(x \oplus W), \label{A6-eq15}\\
    &\mathbb{P}_{f}(P' \oplus x \oplus W) \leq \alpha^{-1} \mathbb{P}_{f}(P') \mathbb{P}_{f}(x \oplus W^*). \label{A6-eq16} 
\end{align}
Substitute \eqref{A6-eq14}-\eqref{A6-eq16} into \eqref{A6-eq13}, we have
\begin{align*}
    \mathbb{P}_{\mathcal{D}}(W|P' \oplus x) - \mathbb{P}_{\mathcal{D}}(W^*|P' \oplus x) \geq \frac{\sum_{f \in \mathcal{D}} Pr(f|\mathcal{D}) \mathbb{P}_{f}(P') ( \alpha^2 \mathbb{P}_{f}(x \oplus W) - \alpha^{-2} \mathbb{P}_{f}(x \oplus W^*) ) }{\sum_{f \in \mathcal{D}} Pr(f|\mathcal{D}) \mathbb{P}_{f}(P')  \mathbb{P}_{f}(x)}.
\end{align*}
Define 
\begin{align*}
    &A := Pr(f^*|\mathcal{D}) \mathbb{P}_{f^*}(P') ( \alpha^2 \mathbb{P}_{f^*}(x \oplus W) - \alpha^{-2} \mathbb{P}_{f^*}(x \oplus W^*) ), \\
    &B := \sum_{f \in \mathcal{D}, f \ne f^*} Pr(f|\mathcal{D}) \mathbb{P}_{f}(P') ( \alpha^2 \mathbb{P}_{f}(x \oplus W) - \alpha^{-2} \mathbb{P}_{f}(x \oplus W^*) ), \\
    &C := Pr(f^*|\mathcal{D}) \mathbb{P}_{f^*}(P')  \mathbb{P}_{f^*}(x), \\
    &D := \sum_{f \in \mathcal{D}, f \ne f^*} Pr(f|\mathcal{D}) \mathbb{P}_{f}(P')  \mathbb{P}_{f}(x),
\end{align*}
we then have
\begin{align*}
    \mathbb{P}_{\mathcal{D}}(W|P' \oplus x) - \mathbb{P}_{\mathcal{D}}(W^*|P' \oplus x) \geq \frac{A}{C+D} + \frac{B}{C+D}.
\end{align*}

Next, we derive upper bounds of $|\frac{B}{C}|$ and $|\frac{D}{C}|$. We have
\begin{align*}
    \bigg| \frac{B}{C} \bigg| = \bigg| \frac{\sum_{f \in \mathcal{D}, f \ne f^*} Pr(f|\mathcal{D}) \mathbb{P}_{f}(P') ( \alpha^2 \mathbb{P}_{f}(x \oplus W) - \alpha^{-2} \mathbb{P}_{f}(x \oplus W^*) )}{Pr(f^*|\mathcal{D}) \mathbb{P}_{f^*}(P')  \mathbb{P}_{f^*}(x)} \bigg| \leq \sum_{f \in \mathcal{D}, f \ne f^*} \bigg| \frac{Pr(f|\mathcal{D}) \mathbb{P}_{f}(P') ( \alpha^2 \mathbb{P}_{f}(x \oplus W) - \alpha^{-2} \mathbb{P}_{f}(x \oplus W^*) )}{Pr(f^*|\mathcal{D}) \mathbb{P}_{f^*}(P')  \mathbb{P}_{f^*}(x)} \bigg|.
\end{align*}
Since we have 
\begin{align*}
     \alpha^2 \mathbb{P}_{f}(x \oplus W) - \alpha^{-2} \mathbb{P}_{f}(x \oplus W^*) \leq 1 \leq \alpha^{-2},
\end{align*}
we further bound $|\frac{B}{C}|$ as follows:
\begin{align*}
    \bigg| \frac{B}{C} \bigg| \leq \sum_{f \in \mathcal{D}, f \ne f^*} \frac{Pr(f|\mathcal{D})}{Pr(f|\mathcal{D^*})} \cdot \frac{\mathbb{P}_{f}(P')}{\mathbb{P}_{f^*}(P')} \cdot \alpha^{-2} \cdot \frac{1}{\mathbb{P}_{f^*}(x)} \leq \sum_{f \in \mathcal{D}, f \ne f^*} \frac{1}{\gamma} \cdot \frac{\mathbb{P}_{f}(P')}{\mathbb{P}_{f^*}(P')} \cdot \alpha^{-2} \cdot \beta^{-\ell}
\end{align*}
due to $Pr(f|\mathcal{D^*}) > \beta^\ell$ and $Pr(f|\mathcal{D^*}) \geq \gamma$. According to Lemma~\ref{A6-L1}, as long as $\frac{\mathbb{P}_{f}(P')}{\mathbb{P}_{f^*}(P')} \leq \frac{\mathbb{P}_{\mathcal{D^*}}(W|x) - \mathbb{P}_{\mathcal{D^*}}(\hat{W}|x)}{5\alpha^{-2}\beta^{-\ell}\gamma^{-1}} := \mu$ and the $M$ is properly choosen, we have $\bigg| \frac{B}{C} \bigg| \leq \frac{1}{5} (\mathbb{P}_{\mathcal{D^*}}(W|x) - \mathbb{P}_{\mathcal{D^*}}(\hat{W}|x))$ with probability at least $1-q$. Similarly, we can bound  $|\frac{D}{C}| < \frac{1}{4}$.

Since $C$ and $D$ are non-negative, we have
\begin{align*}
    \bigg| \frac{A}{C+D} - \frac{A}{C} \bigg| = \bigg| \frac{AD}{C^2+CD} \bigg| \leq \bigg| \frac{AD}{C^2} \bigg| =  \bigg| \frac{A}{C} \bigg| \cdot \bigg| \frac{D}{C} \bigg|,
\end{align*}
implying
\begin{align}\label{A6-eq17} 
    \frac{A}{C+D} \geq \frac{A}{C} - \bigg| \frac{A}{C} \bigg| \cdot \bigg| \frac{D}{C} \bigg| \geq \frac{A}{C} \bigg( 1 - \bigg| \frac{D}{C} \bigg|\bigg) \geq \frac{3}{4} \frac{A}{C}.
\end{align}
Similarly, we can bound 
\begin{align}\label{A6-eq18} 
    \frac{B}{C+D} \geq \frac{B}{C} - \bigg| \frac{B}{C} \bigg| \bigg| \frac{D}{C} \bigg| \geq - \bigg| \frac{B}{C} \bigg| \bigg(1 + \bigg| \frac{D}{C} \bigg|\bigg) \geq -\frac{5}{4} \bigg| \frac{B}{C} \bigg|.
\end{align}

To bound $\frac{A}{C}$, by definition we have
\begin{align}\label{A6-eq19} 
    \frac{A}{C} = \mathbb{P}_{\mathcal{D^*}}(W|x) - \mathbb{P}_{\mathcal{D^*}}(\hat{W}|x) + (\alpha^2-1) \mathbb{P}_{f}(W|x) + (\alpha^{-2}-1)\mathbb{P}_{f}(W^*|x) > \mathbb{P}_{\mathcal{D^*}}(W|x) - \mathbb{P}_{\mathcal{D^*}}(\hat{W}|x) -1 + \alpha^2.
\end{align}
According to \eqref{A6-eq17}-\eqref{A6-eq19}, we have
\begin{align*}
    \mathbb{P}_{\mathcal{D}}(W|P' \oplus x) - \mathbb{P}_{\mathcal{D}}(W^*|P' \oplus x) > \frac{3}{4} (\mathbb{P}_{\mathcal{D^*}}(W|x) - \mathbb{P}_{\mathcal{D^*}}(\hat{W}|x) -1 + \alpha^2) - \frac{5}{4} \cdot \frac{1}{5} (\mathbb{P}_{\mathcal{D^*}}(W|x) - \mathbb{P}_{\mathcal{D^*}}(\hat{W}|x)) > \frac{1}{2} (\mathbb{P}_{\mathcal{D^*}}(W|x) - \mathbb{P}_{\mathcal{D^*}}(\hat{W}|x)) + \alpha^2 -1.
\end{align*}
We then finish the proof.

\section{Proof of Theorem~\ref{T6.3}}\label{A7}

We first prove the upper bound of ICL prediction loss. By choosing $\mu = \frac{\mathbb{P}_{\mathcal{D^*}}(W|x) - \mathbb{P}_{\mathcal{D^*}}(\hat{W}|x)}{(1-\frac{c}{2})^{-1}\alpha^{-2}\beta^{-\ell}\gamma^{-1}}$ and $M \geq \max\bigg\{\frac{-(\ln q) (16\ell^2)(\ln^2\beta)}{KL^2(\mathbb{P}_{f}, \mathbb{P}_{f^*})}, \frac{-2\ln\mu}{\min_{\phi} KL(\mathbb{P}_{f}, \mathbb{P}_{f^*})+8\ln(\alpha \beta)}\bigg\}$, according to Lemma~\ref{L6-2}, we have
\begin{align*}
    \mathbb{P}_{\mathcal{D}}(W|P' \oplus x) - \mathbb{P}_{\mathcal{D}}(W^*|P' \oplus x) > (1-c) (\mathbb{P}_{\mathcal{D^*}}(W|x) - \mathbb{P}_{\mathcal{D^*}}(\hat{W}|x)) + \alpha^{2} -1. 
\end{align*}

Consider that $\epsilon = \frac{2\Delta_{pre}}{c(1-c)}$, where $\Delta_{pre} > \max_{x, W \in \Omega} |\mathbb{P}_{f}(W|x) - \mathbb{P}_{\theta}(W|x)|$ denote the maximum difference between the LLM's pre-trained distribution $\mathbb{P}_{\theta}(\cdot|\cdot)$ and any mapping $f$$\in$$\mathcal{F}$.  If $\mathbb{P}_{\mathcal{D}}(W|P' \oplus x) - \mathbb{P}_{\mathcal{D}}(W^*|P' \oplus x) > \epsilon$, we have
\begin{align*}
    \mathbb{P}_{\mathcal{D}}(W|P' \oplus x) - \mathbb{P}_{\mathcal{D}}(W^*|P' \oplus x) &> (1-c) (\mathbb{P}_{\mathcal{D^*}}(W|x) - \mathbb{P}_{\mathcal{D^*}}(\hat{W}|x)) + \alpha^{2} -1 \\
    &> (1-c) (\mathbb{P}_{\mathcal{D^*}}(W|x) - \mathbb{P}_{\mathcal{D^*}}(\hat{W}|x)) - (1-c)^2 \Delta_{\mathcal{D^*}} \\
    &\geq (1-c) (\mathbb{P}_{\mathcal{D^*}}(W|x) - \mathbb{P}_{\mathcal{D^*}}(\hat{W}|x)) - (1-c)^2 (\mathbb{P}_{\mathcal{D^*}}(W|x) - \mathbb{P}_{\mathcal{D^*}}(\hat{W}|x)) \\
    &= (1-c)c (\mathbb{P}_{\mathcal{D^*}}(W|x) - \mathbb{P}_{\mathcal{D^*}}(\hat{W}|x)) > 2\Delta_{pre},
\end{align*}
which implies that the ICL prediction is exactly the same as the ground truth and there incurs no ICL prediction loss. If $\mathbb{P}_{\mathcal{D}}(W|P' \oplus x) - \mathbb{P}_{\mathcal{D}}(W^*|P' \oplus x) \leq \epsilon$, the mismatching probability of our ICL prediction is less than $\epsilon = \frac{2\Delta_{pre}}{c(1-c)}$, indicating that our ICL prediction loss to BER is bounded by $\epsilon$. Note that $M \geq \max\bigg\{\frac{-(\ln q) (16\ell^2)(\ln^2\beta)}{KL^2(\mathbb{P}_{f}, \mathbb{P}_{f^*})}, \frac{-2\ln(\frac{\epsilon}{2(1-\frac{c}{2})^{-1}\alpha^{-2}\beta^{-T}\gamma^{-1}})}{\min_{f} KL(\mathbb{P}_{f}, \mathbb{P}_{f^*})+8\ln(\alpha \beta)}\bigg\}$ implies $M \geq \max\bigg\{\frac{-(\ln q) (16\ell^2)(\ln^2\beta)}{KL^2(\mathbb{P}_{f}, \mathbb{P}_{f^*})}, \frac{-2\ln\mu}{\min_{\phi} KL(\mathbb{P}_{f}, \mathbb{P}_{f^*})+8\ln(\alpha \beta)}\bigg\}$ for the first case of $\mathbb{P}_{\mathcal{D}}(W|P' \oplus x) - \mathbb{P}_{\mathcal{D}}(W^*|P' \oplus x) > \epsilon$.

Next, we prove our throughput loss bound. We have the ICL prediction loss $ Pr( \mathbb{E}[\mathbf{1}_{(\hat{W} \ne W)}] \leq \epsilon + \text{BER}) \geq 1-q$, where $\text{BER}$ stands for error rate of the Bayes optimal classifier. With probability of at least $1-q$,
We have 
\begin{align*}
 \mathbb{E}[U(W) - U(\hat{W})] 
 =&\mathbb{E}[|U(W) - U(\hat{W}|] \\
 \leq&  \mathbb{E} \bigg[\frac{T_P \bar{N}}{8T_\sigma} \cdot|W - \hat{W}|\bigg] \\
 \leq& \frac{T_P \bar{N}}{8T_\sigma} \cdot \bigg(\mathbb{E} (W-\hat{W})^2 \bigg)^\frac{1}{2} \\
 \leq& \frac{T_P \bar{N}}{8T_\sigma} \cdot \bigg( \mathbb{E} [\mathbf{1}_{(W \ne \hat{W})} \bar{W}^2] \bigg)^\frac{1}{2} \\
 \leq& \frac{T_P \bar{N}}{8T_\sigma} \cdot \bigg(\bigg(\epsilon + \text{BER}\bigg) \bar{W}^2\bigg)^\frac{1}{2} \\
 =& \frac{T_P \bar{N}}{8T_\sigma} \cdot \bigg(\epsilon  +  \cdot \text{BER}\bigg)^\frac{1}{2}\bar{W} = \mathcal{O}\bigg(\frac{T_P \bar{N}}{8T_\sigma}\epsilon^\frac{1}{2} \bar{W}\bigg),
\end{align*}
where the first inequality holds due to Lemma~\ref{L5.4} and the second holds due to Jensen's inequality. We then finish the proof.
\end{document}